\documentclass[nohyperref]{article}

\usepackage[accepted]{icml2024}
 
\usepackage{hyperref}
\definecolor{citeColor}{RGB}{0,20,115}
\hypersetup{colorlinks,linkcolor={citeColor},citecolor={citeColor},urlcolor={citeColor}}  

\usepackage[textsize=tiny]{todonotes}
\usepackage{xspace}

\usepackage{subfigure}
\usepackage{booktabs} 

\usepackage{amsmath}
\usepackage{amssymb}
\usepackage{mathtools}
\usepackage{amsthm}
\usepackage{algorithm}
\usepackage{algorithmic}
\usepackage{amsfonts}
\usepackage[utf8]{inputenc} 
\usepackage[T1]{fontenc}    
\usepackage{hyperref}       
\usepackage{url}            
\usepackage{booktabs}       
\usepackage{amsfonts}       
\usepackage{nicefrac}       
\usepackage{microtype}      
\usepackage{xcolor}         

\usepackage{dsfont}
\usepackage{array}
\usepackage{algorithm}
\usepackage{algorithmic}
\usepackage[algo2e,ruled,noend,vlined,linesnumbered]{algorithm2e}
\usepackage{amsthm}
\usepackage{amsfonts}
\usepackage{enumerate}
\usepackage{subfigure}
\usepackage{stfloats}
\usepackage{float}
\usepackage{bm}
\usepackage{bbm}
\usepackage{tabularx}
\usepackage{enumitem}
\usepackage{multirow}
\usepackage{xcolor, colortbl}
\usepackage{nicefrac}
\usepackage{booktabs}
\usepackage{bbding}
\usepackage{caption}
\usepackage{graphicx}
\usepackage{bbding}
\usepackage{wrapfig}
\usepackage{lipsum}
\usepackage{amssymb}
\usepackage{subfigure}
\usepackage{multibib}
\usepackage{tcolorbox}

\theoremstyle{plain}
\newtheorem{condition}{Condition}
\newtheorem{theorem}{Theorem}
\newtheorem{definition}[theorem]{Definition}
\def \cH {\mathcal{H}}
\usepackage[textsize=tiny]{todonotes}
\definecolor{darkblue}{rgb}{0, 0.12, 0.55}
\definecolor{darkgreen}{rgb}{0, 0.55, 0.12}
\definecolor{darkred}{rgb}{0.55, 0.12,  0}
\definecolor{Gray}{gray}{0.9}

\newcommand{\eg}{\textit{e.g.}}
\newcommand{\ie}{\textit{i.e.}}
\newcommand{\myPara}[1]{\vspace{.05in}\noindent\textbf{#1}}
\usepackage{etoc}
\etocdepthtag.toc{mtchapter}
\etocsettagdepth{mtchapter}{subsection}
\etocsettagdepth{mtappendix}{none}
\icmltitlerunning{Refined Coreset Selection: Towards Minimal Coreset Size under Model Performance Constraints}

\begin{document}
	
	
	\twocolumn[
	\icmltitle{Refined Coreset Selection: Towards Minimal Coreset Size under Model Performance Constraints}

	
	
	\icmlsetsymbol{equal}{*}
	
	\begin{icmlauthorlist}
\icmlauthor{Xiaobo Xia}{USYD}
\icmlauthor{Jiale Liu}{XDU}
\icmlauthor{Shaokun Zhang}{PSU}
\icmlauthor{Qingyun Wu}{PSU}
\icmlauthor{Hongxin Wei}{SUSTech}
\icmlauthor{Tongliang Liu}{USYD}

\end{icmlauthorlist}

\icmlaffiliation{USYD}{The University of Sydney, Australia.}
\icmlaffiliation{XDU}{Xidian University, China.}
\icmlaffiliation{PSU}{Penn State University, USA.}
\icmlaffiliation{SUSTech}{Southern University of Science and Technology, China}
\icmlcorrespondingauthor{Tongliang Liu}{tongliang.liu@sydney.edu.au}

	
	\icmlkeywords{Machine Learning, ICML}
	
	\vskip 0.3in
	]
	
	
	
	\printAffiliationsAndNotice{}  
	
	\begin{abstract}
		Coreset selection is powerful in reducing computational costs and accelerating data processing for deep learning algorithms. It strives to identify a small subset from large-scale data, so that training only on the subset practically performs on par with full data. Practitioners regularly desire to identify the smallest possible coreset in realistic scenes while maintaining comparable model performance, to minimize costs and maximize acceleration. Motivated by this desideratum, for the first time, we pose the problem of refined coreset selection, in which the minimal coreset size under model performance constraints is explored. Moreover, to address this problem, we propose an innovative method, which maintains optimization priority order over the model performance and coreset size, and efficiently optimizes them in the coreset selection procedure. Theoretically, we provide the convergence guarantee of the proposed method. Empirically, extensive experiments confirm its superiority compared with previous strategies, often yielding better model performance with smaller coreset sizes. The code is attached in the supplementary material for the reproducibility of results.
	\end{abstract}

	\section{Introduction}
\label{sec: introduction}

Deep learning has made tremendous strides in recent decades, powered by ever-expanding datasets that comprise millions of examples~\citep{radford2018improving,brown2020language,kirillov2023segany,li2022selective}. At such scales, both data storage and model training become burdensome, and are always unaffordable by startups or non-profit organizations~\citep{zhao2021dataset,liu2022dataset,xia2023moderate}. Hence, there are intense demands for lowering the data scale and improving the data efficiency of deep learning techniques~\citep{sorscher2022beyond,zhao2023dataset,zhao2021dataset,deng2022remember,xie2023data}. 

Coreset selection has been confirmed as a natural and efficacious strategy to satisfy the aforenoted demands~\citep{pooladzandi2022adaptive,feldman2020neural,mirzasoleiman2020coresets,he2023large,lin2023optimal}. This strategy typically involves selecting a small subset (known as a coreset) from massive data. The objective of the selection is that training on the subset can achieve comparable performance to that on the full data. In pursuit of this objective, by first predetermining and fixing the coreset size per request, previous works competed to propose more advanced coreset selection algorithms that better meet the objective~\citep{toneva2019empirical,borsos2020coresets}. Clearly, these works are applicable in the scenario where practitioners have a specific requirement of the coreset size, since subsequent coreset selection is based on it.

In this paper, we go beyond the above scenario and discuss a more general problem about coreset selection, which is named \underline{r}efined \underline{c}oreset \underline{s}election~({RCS}). Specifically, in this problem, we still hold the objective of prior coreset selection, on which the coreset should practically perform on par with full data. Distinctively, we are also concerned about the objective of the coreset size. That is, under the premise of comparable performance achieved by the coreset, its size should be as small as possible for better data efficiency. 

The RCS problem shares a similar philosophy with numerous problems in other domains, which tends to go further on other objectives besides the primary objective~\citep{bommert2017multicriteria,gonzalez2021lexicographic,abdolshah2019multi}. Also, it is much in line with the needs of practitioners. For instance, in lots of cases, we actually do not have a clear and fixed requirement for the coreset size. Instead, if model performance with the coreset can be satisfactory, we desire to further minimize storage and training consumption and are interested in the lowest cost of hardware when utilizing the coreset. This matches the minimal coreset size under model performance constraints. 

To address the RCS problem, we present a new method that formulates RCS as cardinality-constrained bilevel optimization with priority order over multiple objectives. Specifically, we first rigorously formalize the priority order as lexicographic preferences~\citep{fishburn1975axioms,zhang2023targeted}. This helps specify a clear optimization target across multiple objectives, where the model performance is primary and coreset size is secondary. Afterward, with a network trained in the inner loop of bilevel optimization, coreset selection is performed in the outer loop, by using pairwise comparisons between constructed coresets. The pairwise comparisons are supported by lexicographic relations defined for RCS, which proceed toward premium coresets under the lexicographic structure over objectives. 

\subsection{Contributions} 
(1). Conceptually, we surpass the traditional coreset selection paradigm and propose the problem of refined coreset selection~(RCS). The problem is realistic, challenging, and under-explored. The solution for it is non-trivial~(\textit{c.f.}, \S\ref{sec:2.1}). 

(2). Technically, we propose an advanced method to handle RCS, in which lexicographic bilevel coreset selection is framed. We also discuss implementation tricks to speed up the coreset selection in our method. Moreover, theoretical analysis is provided to guarantee the convergence of the proposed method.

(3). Empirically, extensive evaluations are presented on F-MNIST, SVHN, CIFAR-10, and ImageNet-1k. We demonstrate the utility of the proposed method in tackling RCS. Besides, compared with previous efforts in coreset selection, we illustrate that in many situations, our method can reach competitive model performance with a smaller coreset size, or better model performance with the same coreset size.

	
	\subsection{Related Literature}
Coreset selection has gained much interest from the research community~\citep{huggins2016coresets,huang2018epsilon,braverman2022power,qin2023infobatch,park2022active,zheng2023coveragecentric}. The algorithms of coreset selection are generally divided into two groups. In the first group, the methods design a series of score criteria and sort data points based on the criteria. Afterwards, the data points with smaller or larger scores are selected into the coreset. The score criteria include margin separation~\citep{har2007maximum}, gradient norms~\citep{paul2021deep}, distances to class centers~\citep{sorscher2022beyond,xia2023moderate}, influence function scores~\citep{pooladzandi2022adaptive,yang2023dataset}, \textit{etc}. As a comparison, in the second group, the methods do not design any specific score criteria~\citep{feldman2011unified,lucic2017training,huang2023near}. The coreset is commonly constructed in an optimization manner to satisfy an approximation error~\citep{huang2018epsilon}. Compared with the methods in the first group, the methods in the second group often enjoy more promising theoretical properties and guarantees~\citep{huang2018epsilon,huang2023near}. 

Recently, due to the power to handle hierarchical decision-making problems, bilevel optimization~\citep{bard2013practical,eichfelder2010multiobjective,sinha2017review} is introduced to improve the methods in the second group~\citep{borsos2020coresets}. Specifically, the motivation for bilevel coreset selection is that the only thing we really care about is the performance of the model trained on the coreset, instead of a small approximation error for the loss function in the whole parameter space~\citep{zhou2022probabilistic}. Therefore, the approximation error is discarded in optimization. We choose to evaluate the performance~(\eg, the loss) of parameters achieved by training with the selected coreset, on full data. The evaluations are used to guide subsequent coreset selection. The proposed method in this paper is inspired by bilevel coreset selection. Nevertheless, there are prioritized multiple objectives when evaluating performance, which is more challenging both intuitively and technically.

Bilevel multi-objective optimization~\citep{deb2010efficient,sinha2015towards,gu2023min} imposes multiple objectives in each loop of a bilevel optimization problem. Our algorithm design is related to bilevel multi-objective optimization~\citep{deb2010efficient}, in the sense that there are two evaluation objectives in the outer loop of bilevel optimization. However, to the best of our knowledge, there is no study exploring coreset selection with bilevel multi-objective optimization. Therefore, from this perspective, this paper benefits the community in two folds: (1). we investigate coreset selection with bilevel multi-objective optimization and discuss the issues of this paradigm; (2). we present the algorithm of bilevel coreset selection with priority structures to address the issues, which can inspire follow-up research.

	\section{Preliminaries}

\myPara{Notations.} In the sequel, vectors, matrices, and tuples are denoted by bold-faced letters. We use $\|\cdot\|_p$ to denote the $L_p$ norm of vectors or matrices and $\ell(\cdot)$ to denote the cross-entropy loss if there is no confusion. Let $[n]=\{1,\ldots,n\}$. 

\myPara{Problem definition.} We define the problem of RCS as follows. Formally, given a large-scale dataset $\mathcal{D}=\{(\mathbf{x}_i,y_i)\}_{i=1}^n$ with a sample size $n$, where $\mathbf{x}_i$ denotes the instance and $y_i$ denotes the label. The problem of RCS is to find a subset of $\mathcal{D}$ for follow-up tasks, which reduces both storage and training consumption while maintaining the utility. The subset is called the \textit{coreset} that is expected to satisfy two objectives by priority: \textbf{(O1)} the coreset should practically perform on par with full data $\mathcal{D}$; \textbf{(O2)} the sample size of the coreset should be as small as possible. Note that objective \textbf{(O1)} has a higher priority than \textbf{(O2)}, since a smaller coreset size is pointless if the network with this small coreset does not perform satisfactorily.

\myPara{Objective formulations.} We formulate the two optimization objectives that we are concerned with. Without loss of generality, we consider the minimization mode across the paper. The formulation is based on a bilevel optimization framework~\citep{borsos2020coresets,zhou2022probabilistic}. Specifically, the 0-1 masks $\bm{m}\in\{0,1\}^n$ are introduced with $m_i=1$ indicating the data point $(\mathbf{x}_i,y_i)$ is selected into the coreset and otherwise excluded. We use $h(\mathbf{x};\bm{\theta})$ to denote the deep network with the learnable parameters $\bm{\theta}$. The objective (O1) can be formulated as 
\begin{align}
     & f_1(\bm{m}):=\frac{1}{n}\sum_{i=1}^n \ell(h(\mathbf{x}_i;\bm{\theta}(\bm{m})),y_i), \\\nonumber
     & \text{s.t.} ~~\bm{\theta}(\bm{m})\in\arg\min_{\bm{\theta}}\mathcal{L}(\bm{m},\bm{\theta}),
\end{align}
where $\bm{\theta}(\bm{m})$ denotes the network parameters obtained by training the network to converge on the selected coreset with mask $\bm{m}$. That $\mathcal{L}(\bm{m},\bm{\theta})$ represents the loss on the selected coreset with $\mathcal{L}(\bm{m},\bm{\theta})=\frac{1}{\|\bm{m}\|_0}\sum_{i=1}^n m_i\ell(h(\mathbf{x}_i;\bm{\theta}),y_i)$. The intuition of (O1) is that a good coreset ensures optimizing on $\mathcal{L}({\bm{m},\bm{\theta}})$ over $\bm{\theta}$ yields good solutions when evaluated on $f_1(\bm{m})$~\citep{borsos2020coresets}. Also, we define the objective (O2) as 
\begin{equation}
    f_2(\bm{m}):=\|\bm{m}\|_0, 
\end{equation}
which explicitly controls the coreset size using $L_0$ norm. In this work, we aim to minimize $f_1(\bm{m})$ and $f_2(\bm{m})$ in order of priority, where $f_1(\bm{m})$ is primary and $f_2(\bm{m})$ is secondary. That $f_2(\bm{m})$ should be optimized under the premise of $f_1(\bm{m})$. 

\subsection{RCS Solutions are Non-trivial}\label{sec:2.1}

\begin{figure}[!t]
	\centering
	\subfigure[]{
		\begin{minipage}[b]{0.22\textwidth}
			\includegraphics[width=1.05\textwidth]{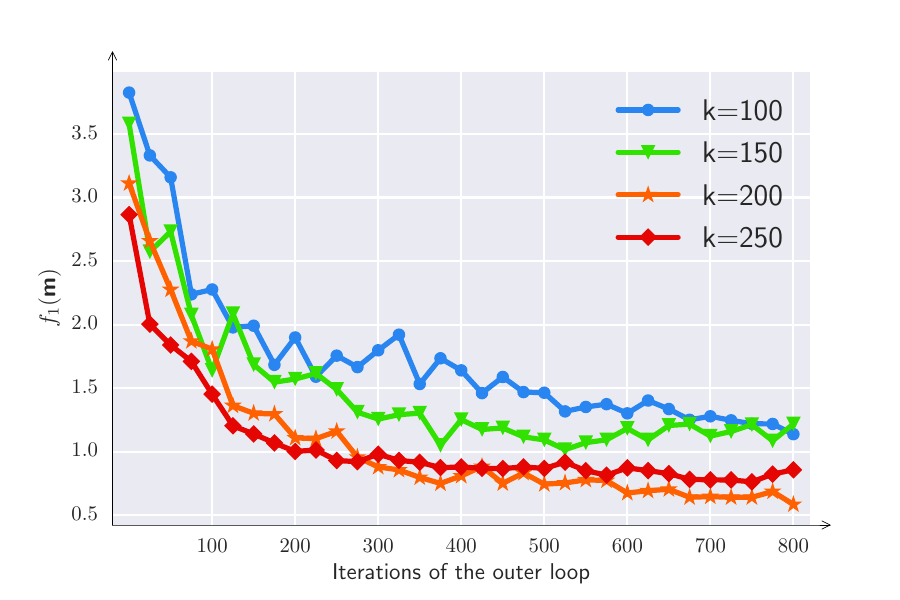} 
		\end{minipage}
		\label{fig:a}
	}
    	\subfigure[]{
    		\begin{minipage}[b]{0.22\textwidth}
   		 	\includegraphics[width=1.05\textwidth]{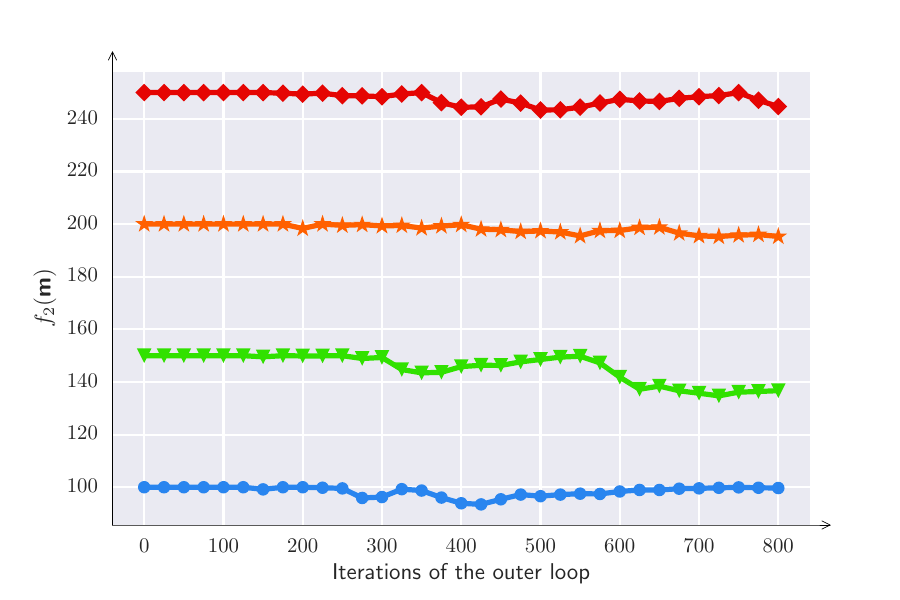}
    		\end{minipage}
		\label{fig:b}
    	}
	\subfigure[]{
		\begin{minipage}[b]{0.22\textwidth}
			\includegraphics[width=1.05\textwidth]{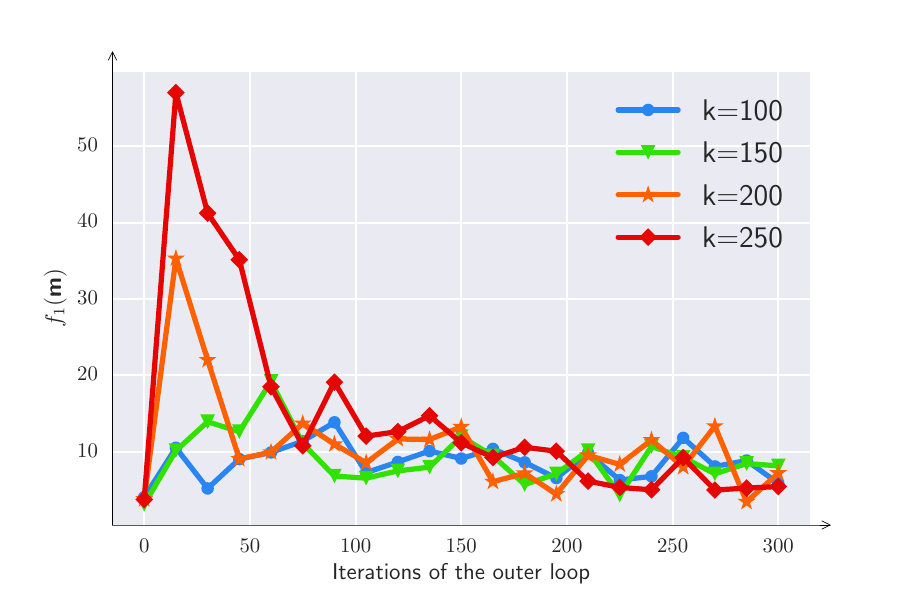} 
		\end{minipage}
		\label{fig:c}
	}
    	\subfigure[]{
    		\begin{minipage}[b]{0.22\textwidth}
		 	\includegraphics[width=1.05\textwidth]{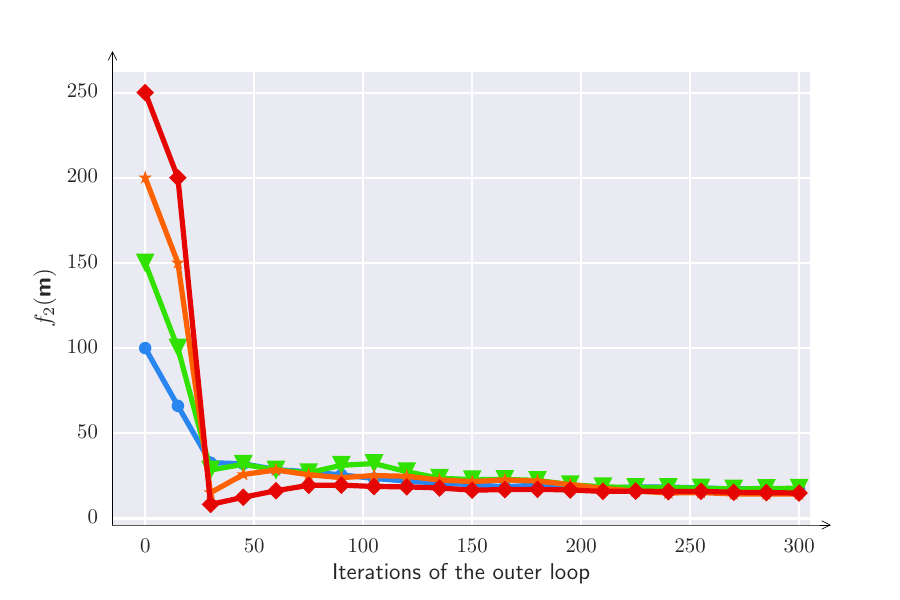}
    		\end{minipage}
		\label{fig:d}
    	}
    \vspace{-3pt}
    \caption{Illustrations of phenomena of several trivial solutions discussed in \S\ref{sec:2.1}. The experiment is based on~\citep{zhou2022probabilistic}. The setup is provided in Appendix~\ref{supp:exp_fig1}. Here, $k$ denotes the predefined coreset size \textit{before} optimization. \textbf{(a)} $f_1(\bm{m})$ vs. outer iterations with (\ref{eq:cs_bi}); \textbf{(b)} $f_2(\bm{m})$ vs. outer iterations with (\ref{eq:cs_bi}); \textbf{(c)} $f_1(\bm{m})$ vs. outer iterations with (\ref{eq:cs_bi_mpo}); \textbf{(d)} $f_2(\bm{m})$ vs. outer iterations with (\ref{eq:cs_bi_mpo}). }
    \vspace{-5pt}
\end{figure}

Solving RCS is non-trivial since previous methods on coreset selection can not be applied directly. Moreover, simple modifications to these methods may not be sufficient to achieve good solutions. For the attempt of direct applications, it is somewhat easy to know this is infeasible. Most works fix the coreset size for coreset selection~\citep{paul2021deep,xia2023moderate,sorscher2022beyond,toneva2019empirical}. Also, the methods~\citep{borsos2020coresets,zhou2022probabilistic} in bilevel optimization specify a predefined upper bound of the coreset size, and only consider the objective (O1) in optimization:
\begin{equation}\label{eq:cs_bi}
    \min_{\bm{m}} f_1(\bm{m}), \ \ \text{s.t.}\ \ \bm{\theta}(\bm{m})\in\arg\min_{\bm{\theta}}\mathcal{L}(\bm{m},\bm{\theta}).
\end{equation}
In (\ref{eq:cs_bi}), the minimization of $f_1(\bm{m})$ is in the outer loop, while the minimization of $\mathcal{L}(\bm{m},\bm{\theta})$ lies in the inner loop. Without optimizations about the coreset size, $f_1(\bm{m})$ can be minimized effectively (see Figure~\ref{fig:a}). As a comparison, the coreset size remains close to the predefined one~(see Figure~\ref{fig:b}), which is not our desideratum in RCS. 

In an attempt to modify previous methods to tackle RCS, we discuss two simple-to-conceive cases. To begin with, for the methods that fix the coreset size for subsequent coreset selection, we can borrow them to run many experiment attempts under different coreset sizes. The attempts with comparable $f_1(\bm{m})$ and small $f_2(\bm{m})$ can be employed as a solution. However, this way needs expert assistance for lower attempt budgets~\citep{yao2018taking}. Also, its performance is not very competitive (see evaluations in \S\ref{sec:experiments}).

In addition, for the methods in bilevel coreset selection, by introducing the objective (O2) to (\ref{eq:cs_bi}), we can minimize two objectives in the form of a weighted combination: 
\begin{equation}\label{eq:cs_bi_mpo}
     \min_{\bm{m}} (1-\lambda)f_1(\bm{m})+\lambda f_2(\bm{m}), \ \ \text{s.t.} \ \ \bm{\theta}(\bm{m})\in\arg\min_{\bm{\theta}}\mathcal{L}(\bm{m},\bm{\theta}),
\end{equation}
where $\lambda\in(0,1)$ is a hyper-parameter to balance the two objectives in (\ref{eq:cs_bi_mpo}). First, intuitively, as $f_2(\bm{m})$ has lower priority than $f_1(\bm{m})$ in RCS, we can tune a smaller weight for $f_2(\bm{m})$, \ie, $\lambda<1/2$. Unfortunately, it is intractable, since the two objectives have different magnitudes that are related to data, networks, optimization algorithms, and specific tasks simultaneously~\citep{gong2021automatic}. Second, if $f_1(\bm{m})$ and $f_2(\bm{m})$ share the same weights, \ie, $\lambda=1/2$, optimization does not implicitly favor $f_1(\bm{m})$. Instead, the minimization of $f_2(\bm{m})$ is salient, where after all iterations $f_2(\bm{m})$ is too small and $f_1(\bm{m})$ is still large (see Figures~\ref{fig:c} and \ref{fig:d}). This contradicts our aim in RCS, since satisfactory network performance achieved by the coreset has a higher priority order. With the work~\citep{zhou2022probabilistic}, to explain the experimental phenomenon, we provide the analysis with gradient norms of objectives. The gradient derivations are presented mathematically
in Appendix~\ref{supp:gradients}.

Therefore, based on the above discussions, we can know that RCS solutions are non-trivial. This demonstrates the urgency of developing more advanced algorithms.
	
	\section{Methodology}
\subsection{Lexicographic Bilevel Coreset Selection}
Although both (O1) and (O2) are optimization objectives we care about, in optimization, there is a priority order between them. As analyzed above, (O1) has a higher
priority than (O2), since a smaller coreset size is meaningless if the network with such a small coreset does
not perform satisfactorily. We formalize a general notion of priority order rigorously as a lexicographic preference~\citep{fishburn1975axioms} over two objectives. The general notion helps specify a clear optimization target across multiple objectives before
optimization and avoids manual post hoc selection.

Specifically, we use the order list ${F}(\bm{m})=[f_1(\bm{m}),f_2(\bm{m})]$ to represent the objectives with a lexicographic structure,  in which $f_1$ is the objective with higher priority and $f_2$ is the one with lower priority. The optimization of $f_2$ only matters on the condition that the more important objective $ f_1$ is well-optimized. Afterward, our lexicographic bilevel coreset selection can be formulated as 
\begin{equation}\label{eq:lexi}
    \vec{\min}_{\bm{m}\in\mathcal{M}} F(\bm{m}), \ \ \text{s.t.}\ \ \bm{\theta}(\bm{m})\in\arg\min_{\bm{\theta}}\mathcal{L}(\bm{m},\bm{\theta}),
\end{equation}
where $\vec{\min}$ represents the lexicographic optimization
procedure over the ordered list $F(\bm{m})$ \citep{zhang2023targeted} and  $\mathcal{M}$ denotes the search space of the mask $\bm{m}$. It is worth mentioning that the outer loop is not reflected by a single utility function enclosing both $f_1$ and $f_2$. The reason is that, mathematically, it is impossible to construct a single utility function that represents lexicographic preferences as weighted objectives~(\textit{c.f.},~\citep{shi2020utility}).

\myPara{Remark~1.} Compared with the trivial solution, \ie, the form of weighted combination in (\ref{eq:cs_bi_mpo}), our lexicographic bilevel
coreset selection enjoys several advantages. (i). Our method does not need to determine the combination weight, which is helpful for optimization when the two objectives are of different scales. (ii). Our method can reach Pareto optimality, where the weighted combination falls short~\citep{zhang2023targeted}. (iii). When a research problem has a clear hierarchy of objectives where some objectives are definitively more important than others, lexicographic preference aligns with the decision-making process more naturally than the weighted combination. These advantages explain why the proposed way is better than the trivial method in solving the trade-off between $f_1$ and $f_2$.

\begin{algorithm}[!t]
1: {\bfseries Require:} a network $\bm{\theta}$, a dataset $\mathcal{D}$, a predefined size $k$, and voluntary performance compromise $\epsilon$;\\
2: {\bfseries Initialize} masks $\bm{m}$ randomly with $\|\bm{m}\|_0=k$;

\For{\rm{training iteration} $t = 1,2,\dots,T$}{
	
	3: {\bfseries Train} the inner loop with $\mathcal{D}$ to converge satisfies: $\bm{\theta}(\bm{m})\leftarrow\arg\min_{\bm{\theta}}\mathcal{L}(\bm{m},\bm{\theta})$;\\
	4: {\bfseries Update} masks $\bm{m}$ with $\bm{\theta}(\bm{m})$ by lexicographic optimization as discussed in \S\ref{sec:lexico_optimization};

}
5: {\bfseries Output:} masks $\bm{m}$ after all training iterations.
\caption{Lexicographic bilevel coreset selection~(LBCS) for RCS.}

\label{alg:ours}
\end{algorithm}

\subsection{Optimization Algorithm}\label{sec:lexico_optimization}
\myPara{Challenges.}  We discuss the optimization details of lexicographic bilevel coreset selection that is formulated in (\ref{eq:lexi}). The optimization of the inner loop is simple by directly minimizing $\mathcal{L}(\bm{m},\bm{\theta})$. It is challenging to optimize the outer loop that has a priority structure. As under lexicographic optimization, it is inaccessible to the gradients of $f_1(\bm{m})$ and $f_2(\bm{m})$ with respect to $\bm{m}$, the methods that require analytic forms of gradients~\citep{gong2021automatic} are inapplicable. Also, it is inefficient to borrow multi-objective optimization methods~\citep{gunantara2018review} to find Pareto frontiers~\citep{lotov2008visualizing}, since the found Pareto frontiers are widespread. Actually, we are only interested in a subset of them in a specific region.

\myPara{Black-box optimization.} Given these considerations, we propose to treat the optimization of the outer loop as a black-box optimization problem and leverage a randomized direct search algorithm to solve it. 
The optimization algorithm only needs a set of binary relations used to compare any two masks with their evaluation values for the two objectives $f_1$ and $f_2$. The evaluation results of different masks are iteratively queried, leading to the best mask to solve the RCS problem. The core of the optimization is lexicographic relations~\citep{zhang2023targeted} that are used to compare the performance of different masks with respect to $F(\bm{m})$. We define the lexicographic relations for RCS below.

\begin{definition}[Lexicographic relations in RCS]\label{def:lexico_relations} With two masks for coreset selection, denoted by $\bm{m}$ and $\bm{m}'$ respectively, the lexicographic relations for RCS are defined as 
\begin{align}
    &F(\bm{m})~\vec{=}~F(\bm{m}')\Leftrightarrow f_i(\bm{m})=f_i(\bm{m}')~~\forall i \in  [2],\\\nonumber
    &F(\bm{m})~\vec{\prec}~F(\bm{m}')\Leftrightarrow\\\nonumber
    &\exists i \in [2]: f_i(\bm{m})<f_i(\bm{m}')\land\left(\forall i'<i, f_{i'}(\bm{m}) = f_{i'}(\bm{m}')\right),\\\nonumber
    &F(\bm{m})~\vec{\preceq}~F(\bm{m}')\Leftrightarrow F(\bm{m})~\vec{=}~F(\bm{m}')\lor F(\bm{m})~\vec{\prec}~F(\bm{m}').
\end{align}
\end{definition}
It should be noted that the lexicographic relation ``$\vec{\preceq}$'' has been verified to be both \textit{reflexive} and \textit{transitive}~\citep{zhang2023targeted}. Therefore, leveraging the defined lexicographic relations, the comparisons between any two feasible masks are always conclusive. The optimal point of the outer loop under lexicographic optimization is any one element in the optimum set $\mathcal{M}^*=\{\bm{m}\in\mathcal{M}^*_2~|~\forall \bm{m}\neq\bm{m}',F(\bm{m})~\vec{\preceq}~F(\bm{m}')\}$. Here $\mathcal{M}_2^*$ is defined recursively as 
\begin{equation}
\fontsize{9}{1}\selectfont
    \begin{aligned}\label{eq:lexico_set}
    &\mathcal{M}_1^*:=\{\bm{m}\in\mathcal{M}~|~f_1(\bm{m})\leq f_1^**(1+\epsilon)\}, f_1^*:=\inf_{\bm{m}\in\mathcal{M}}f_1(\bm{m}),\\\nonumber
    &\mathcal{M}_2^*:=\{\bm{m}\in\mathcal{M}^*_1~|~f_2(\bm{m})\leq f_2^*\}, \ \ \text{and} \ \ f_2^*:=\inf_{\bm{m}\in\mathcal{M}_1^*}f_2(\bm{m}), 
\end{aligned}
\end{equation}
where $\epsilon$ represents the percentage of \textit{voluntary performance compromise} of $f_1(\bm{m})$ to find choices with better performance on $f_2(\bm{m})$. In RCS, it is a non-negative number. 

\myPara{Remark~2}~(On the compromise of $f_1(\bm{m})$). A relatively small compromise of $f_1(\bm{m})$ does not necessarily degrade the model performance by the coreset when generalizing to test data. Instead, the compromise saves $f_1(\bm{m})$ from having to be optimized to the minimum, which \textit{reduces the model overfitting} in coreset selection. This can help the model generalization, especially when training data for coreset selection are polluted, \eg, corrupted by mislabeled data. The previous method such as \citet{zhou2022probabilistic} did not take the issue of overfitting into consideration, but moved towards the minimum in optimization. Its performance would be degraded in a series of cases. More details and evidence can be found in \S\ref{sec:comparison} and \S\ref{sec:imperfect}.

\myPara{Algorithm flow and tricks for acceleration.} The lexicographic optimization flow of the outer loop of (\ref{eq:lexi}) is provided in Appendix~\ref{supp:lexiflow}. Besides, the overall algorithm flow of the proposed \underline{l}exicographic \underline{b}ilevel \underline{c}oreset \underline{s}election~(\textbf{LBCS})  for RCS is shown in Algorithm~\ref{alg:ours}. The computational consumption of Algorithm~\ref{alg:ours} originates from the model training in the inner loop~(Step~3) and mask updates in the outer loop~(Step~4). To speed up the inner loop, we can first train a model with random masks and then finetune it with other different masks in Step~3. Also, we can employ model sparsity and make the trained model smaller for faster training. To accelerate the outer loop, the mask search space can be narrowed by treating several examples as a group. The examples in the same group share the same mask in coreset selection. These tricks make our method applicable to large-scale datasets. 

\section{Theoretical Analysis}

We begin by introducing notations and notions. Specifically, for an objective function $f$, its infimum value in the search space $\mathcal{M}$ is denoted by $f^*$. We employ $\bm{m}^t$ to represent the mask at the $t$-th iteration generated by our algorithm. That $\{\bm{m}^t\}_{t=0}^T$ denotes the sequence of masks generated by the proposed algorithm upon the step $T$ from the start time~($t=0$). Also, $\psi_t$ represents the probability measure in the step $t$, which is defined on the search space $\mathcal{M}$. In the following, we present \textit{progressable} and \textit{stale moving} conditions to facilitate theoretical analysis of our LBCS. 

\begin{condition}[Progressable condition]
\label{condition:progressable}
LBCS can optimize objectives $f_{1}$ and $f_{2}$ lexicographically. Namely, at any step $t \geq 0 $, the masks $\bm{m}^{t}$ and $\bm{m}^{t+1}$ satisfy:
\begin{equation}
\label{eq6}
\left\{
\begin{aligned}
&f_{1}(\bm{m}^{t+1}) < f_{1}(\bm{m}^{t}) ~~~~ \text{if} ~~ \bm{m}^{t} \notin \mathcal{M}_{1}^{*};\\
&(f_{2}(\bm{m}^{t+1}) < f_{2}(\bm{m}^{t})) \land (\bm{m}^{t+1} \in \mathcal{M}_{1}^{*}) ~~~~ \text{if} ~~ \bm{m}^{t} \in \mathcal{M}_{1}^{*}.
\end{aligned}
\right.
\end{equation}
\end{condition}
\myPara{Remark 3.} According to lexicographic relations used for mask updates (\textit{c.f.}, Line 10 of Algorithm~2 in Appendix~\ref{supp:lexiflow}), Condition~\ref{condition:progressable} holds at all time steps in LBCS.
Specifically, when $f_{1}$ is not well-optimized, LBCS updates the incumbent mask only if the current evaluating mask has a better value on $f_{1}$. On the other hand, when $f_{1}$ reaches the optimal region $\mathcal{M}_{1}^{*}$, LBCS will update the incumbent mask only if the current evaluating mask has a better value on the second objective $f_{2}$, while $f_{1}$ remains in $\mathcal{M}_{1}^{*}$.

\begin{condition}[Stable moving condition]
\label{condition:lower-bound}
\label{condition:}
At any step $t \geq 0$, (i) if $\bm{m}^t \notin \mathcal{M}^*_{1}$, for all possible $\bm{m}^{t}$ in the set $\mathcal{S}_{1} \coloneqq \{ \bm{m}^{t} \in \mathcal{M} | f(\bm{m}^{t}) \leq f(\bm{m}^{0}) \}$, there exists $ \gamma_1>0 $ and $ 0< \eta_1 \leq 1 $ so that the algorithm satisfies:
\begin{equation}
 \fontsize{10}{1}\selectfont
    \psi_{t+1}[f_1(\bm{m}^{t}) - f_1(\bm{m}^{t+1}) > \gamma_1 ~~\text{or}~~ \bm{m}^{t} \in \mathcal{M}_{1}^* ] \geq \eta_1, 
\end{equation}
and (ii) if $\bm{m}^t \in \mathcal{M}^*_{1}$, for all possible $\bm{m}^{t}$ in the set $\mathcal{S}_{2} \coloneqq \{ \bm{m}^{t} \in \mathcal{M} | f(\bm{m}^{t}) \leq f(\bm{m}^{\hat{t}}) \}$, there exists $ \gamma_2>0 $ and $ 0< \eta_2 \leq 1 $ so that the algorithm satisfies:
\begin{equation}
\fontsize{10}{1}\selectfont
\psi_{t+1}[f_2(\bm{m}^{t}) - f_2(\bm{m}^{t+1}) > \gamma_2 ~~\text{or}~~ \bm{m}^{t} \in \mathcal{M}_{2}^*] \geq \eta_2, 
\end{equation}
where $\hat{t}$ represents the earliest time step that the incumbent mask reaches the optimal region in the objective $f_{1}$, \ie, $\hat{t} \coloneqq \min\{t \in \{\bm{m}^t\}_{t=0}^T| \bm{m}^t \in \mathcal{M}_1^{*}\}$.
\end{condition}

\myPara{Remark 4.}~Condition~\ref{condition:lower-bound} is an assumption that applies to both optimization objectives $f_{1}$ and $f_{2}$, the search space $\mathcal{M}$, and the search algorithm. This condition is commonly used in the convergence analysis of local randomized search algorithms \citep{doi:10.1137/S1052623400374495,solis1981minimization}.
In essence, Condition~\ref{condition:lower-bound} imposes an improvement lower bound on each step of the local randomized search algorithm. This ensures that progress is made stably in each step of the algorithm, and is essential for proving convergence to a globally optimal solution.

With these notations, notions, and conditions, we are ready to exhibit the convergence analysis of our LBCS. Notice that the algorithm convergence in the RCS problem differs from the convergence in traditional multiple objective optimization problems~\citep{morales2022survey,karl2022multi}. In RCS, with two optimization objectives $f_1$ and $f_2$, we say an algorithm is converged if (i) the primary objective $f_1$ reaches the optimum considering the user-provided compromise $\epsilon$; (ii) the secondary objective $f_2$ reaches the optimum under that (i) is satisfied. The theoretical result is formally presented below.

\begin{theorem}[$\epsilon$-convergence] \label{convergence_f1}
Under Condition~\ref{condition:progressable} and Condition~\ref{condition:lower-bound} (sufficient conditions), the algorithm is $\epsilon$-convergence in the RCS problem:
\begin{align}
&\mathbbm{P}_{t \to \infty}[f_{2}(\bm{m}^t) \leq f_2^*] = 1\\\nonumber
&\text{s.t.} ~~ f_2^* = \min\limits_{\bm{m} \in \mathcal{M}}\{f_2(\bm{m})| f_1(\bm{m}) \leq f_1^* * (1+\epsilon)\},
\end{align}
where $\mathbbm{P}[f_{2}(\bm{m}^t) \leq f_2^*]$ represents the probability that the mask $\bm{m}^t$ generated at time \(t\) is the converged solution as described above. 
\end{theorem}
\vspace{-5pt}
The proof of Theorem~\ref{convergence_f1} can be checked in Appendix~\ref{proof:convergence}. 
	\section{Experiments}\label{sec:experiments}
\subsection{Preliminary Presentation of Algorithm's Superiority}
\begin{table}[!h]
    \centering
    \scriptsize
    \begin{tabular}{c|c|c|ccc}
    \toprule
    $k$ & Objectives & Initial & $\epsilon=0.2$ & $\epsilon=0.3$ & $\epsilon=0.4$\\\midrule
    \multirow{2}{*}{200}& $f_1(\bm{m})$ & 3.21 &  1.92$\pm$0.33 & 2.26$\pm$0.35 & 2.48$\pm$0.30\\
    & $f_2(\bm{m})$ & 200 & 190.7$\pm$3.9 & 185.0$\pm$4.6 & 175.5$\pm$7.7\\\midrule
    \multirow{2}{*}{400}& $f_1(\bm{m})$      & 2.16 & 1.05$\pm$0.26 & 1.29$\pm$0.33 & 1.82$\pm$0.41\\
    & $f_2(\bm{m})$ & 400 & 384.1$\pm$4.4 & 373.0$\pm$6.0 & 366.2$\pm$8.1\\\bottomrule
    \end{tabular}
    \vspace{-2pt}
    \caption{Results~(mean$\pm$std.) to illustrate the utility of our method in optimizing the objectives $f_1(\bm{m})$ and $f_2(\bm{m})$.}
    \label{tab:pre_experiments}
\end{table}
\noindent As discussed, there is no previous study specializing in RCS. We therefore only discuss the results achieved by our method. We show that the proposed method can effectively optimize two objectives $f_1(\bm{m})$~(the network performance achieved by the coreset) and $f_2(\bm{m})$~(the coreset size). We conduct experiments on MNIST-S which is constructed by random sampling 1,000 examples from original MNIST~\citep{lecun1998gradient}. Staying with previous work~\citep{borsos2020coresets}, we use a convolutional neural network stacked with two blocks of convolution, dropout, max-pooling, and ReLU activation. The predefined coreset size $k$ is set to 200 and 400 respectively. The voluntary performance compromise of $f_1(\bm{m})$ denoted by $\epsilon$ varies in 0.2, 0.3, and 0.4. All experiments are repeated 20 times on NVIDIA GTX3090 GPUs with PyTorch. The mean and standard deviation (std.) of results are reported. 

We provide results in Table~\ref{tab:pre_experiments}. First, as can be seen, compared with initialized $f_1(\bm{m})$ and $f_2(\bm{m})$, both achieved $f_1(\bm{m})$ and  $f_2(\bm{m})$ after lexicographic bilevel coreset selection are lower. This demonstrates that our method can construct a high-quality coreset with a size that is smaller than the predefined one. Second, we observe that a larger $\epsilon$ will lead to a smaller $f_2(\bm{m})$ under multiple experiments. The phenomenon justifies our previous statements well. Note that here we stress, in one experiment, that a larger $\epsilon$ does not necessarily produce a larger $f_1(\bm{m})$. It is because we only restrict the upper bound of $f_1(\bm{m})$ by $\epsilon$, but not its exact value~(see (\ref{eq:lexico_set})). Nevertheless, when the number of experiments becomes relatively large, on average, achieved $f_1(\bm{m})$ increases accordingly if we increase $\epsilon$.

\begin{table*}[!t]
    \centering
    \scriptsize
    \setlength{\tabcolsep}{2.5mm}{
    \begin{tabular}{l|c|ccccccc|>{\columncolor{Gray}}c|>{\columncolor{Gray}}c}
    \toprule
     & $k$ &Uniform & EL2N & GraNd &Influential & Moderate & CCS & Probabilistic & LBCS~(ours) & Coreset size (ours) \\\midrule
    \parbox[t]{2mm}{\multirow{4}{*}{\rotatebox{90}{\scriptsize F-MNIST}}}& 1000 & 76.9$\pm$2.5 & 71.8$\pm$2.9 & 70.7$\pm$4.0 & 78.9$\pm$2.0 & 77.0$\pm$0.6 & 76.7$\pm$3.5 & \textbf{80.3$\pm$0.6} & 79.7$\pm$0.7 & \textbf{956.7$\pm$3.5}\\
    &2000 & 80.0$\pm$2.4 & 73.7$\pm$1.6 & 71.7$\pm$2.3 & 80.4$\pm$0.8&80.3$\pm$0.4 & 81.4$\pm$0.6 & 82.6$\pm$0.2 & \textbf{82.8$\pm$0.6} & \textbf{1915.3$\pm$6.6} \\
    &3000 & 81.7$\pm$1.7 & 75.3$\pm$2.3 & 73.3$\pm$1.8 & 81.5$\pm$1.2 & 81.7$\pm$0.5 & 82.6$\pm$1.2 &83.7$\pm$0.9 & \textbf{84.0$\pm$0.6} & \textbf{2831.6$\pm$10.9}\\
    &4000 & 83.0$\pm$1.7 & 77.0$\pm$1.0 & 75.9$\pm$2.1 & 82.4$\pm$1.3 & 82.4$\pm$0.3 & 84.1$\pm$0.6 & 84.2$\pm$0.7 & \textbf{84.5$\pm$0.4} & \textbf{3745.4$\pm$15.6}\\\midrule
    \parbox[t]{2mm}{\multirow{4}{*}{\rotatebox{90}{\scriptsize SVHN}}}& 1000 & 67.1$\pm$3.3 & 56.8$\pm$1.3 & 60.7$\pm$1.1 & 70.3$\pm$0.8 & 68.4$\pm$2.0 & 66.9$\pm$1.9 &67.8$\pm$0.4 & \textbf{70.6$\pm$0.3} & \textbf{970.0$\pm$4.8}\\
    & 2000 & 75.9$\pm$1.0 & 64.8$\pm$0.6 & 67.3$\pm$2.0 & 76.2$\pm$1.3 & 77.9$\pm$0.7 & 77.3$\pm$0.8 & 76.6$\pm$1.3 & \textbf{78.3$\pm$0.7} & \textbf{1902.3$\pm$10.3}\\
    & 3000 & 80.3$\pm$1.2 & 72.1$\pm$2.8 & 75.2$\pm$1.6 & 80.8$\pm$1.5 & 81.8$\pm$0.7 & 81.9$\pm$0.6 & 80.9$\pm$1.1 & \textbf{82.3$\pm$0.7} & \textbf{2712.6$\pm$15.0}\\
    &4000 & 83.9$\pm$0.8 & 75.8$\pm$1.9 & 79.1$\pm$2.4 & 83.6$\pm$1.8 & 83.9$\pm$0.6 & 84.1$\pm$0.3 & 84.3$\pm$1.4 & \textbf{84.6$\pm$0.6} & \textbf{3804.2$\pm$16.4}\\\midrule
    \parbox[t]{2mm}{\multirow{4}{*}{\rotatebox{90}{\scriptsize CIFAR-10}}}& 1000 & 46.9$\pm$1.8 & 36.8$\pm$1.2 & 41.6$\pm$2.0 & 45.7$\pm$1.1 & 48.1$\pm$2.2 & 47.6$\pm$1.6 & 48.2$\pm$0.9 & \textbf{48.3$\pm$1.2} & \textbf{970.4$\pm$2.9}\\
    &2000& 58.1$\pm$2.0 & 47.9$\pm$0.7 & 52.3$\pm$2.4 & 57.7$\pm$1.3 & 58.5$\pm$1.3 & 59.3$\pm$1.4 & 60.1$\pm$0.8 & \textbf{60.4$\pm$1.0} & \textbf{1955.2$\pm$5.3}\\
    &3000 & 65.7$\pm$2.3 & 56.1$\pm$1.9 & 61.9$\pm$1.7 & 67.5$\pm$1.6 & 69.2$\pm$2.6 & 67.6$\pm$1.6 & 68.7$\pm$1.1 & \textbf{69.5$\pm$0.9} & \textbf{2913.8$\pm$9.6}\\
    &4000 & 70.9$\pm$2.5 & 63.0$\pm$2.0 & 67.9$\pm$1.3 & 71.7$\pm$2.4 & \textbf{73.9$\pm$0.4} & 73.0$\pm$0.9 & 73.6$\pm$0.2 & 73.4$\pm$0.5 & \textbf{3736.0$\pm$14.2}\\\bottomrule
    \end{tabular}}
    \caption{Mean and standard deviation of test accuracy (\%) on different benchmarks with various predefined coreset sizes. The best mean test accuracy and optimized coreset size by our method in each case are in \textbf{bold}.}
    \label{tab:exact_results}
\end{table*}

\subsection{Comparison with the Competitors}\label{sec:comparison}

\myPara{Competitors.} Multiple coreset selection methods act as baselines for comparison. To our best knowledge, before that, there was no study working on the RCS problem. Therefore, the baselines are the methods that construct the coreset with a predetermined coreset size, where the size is not further minimized by optimization. Specifically, we employ (i). \textit{Uniform sampling}~(abbreviated as Uniform); (ii). \textit{EL2N}~\citep{paul2021deep}; (iii). \textit{GraNd}~\citep{paul2021deep}; (iv). \textit{Influential coreset}~\citep{yang2023dataset}~(abbreviated as Influential); (v). \textit{Moderate coreset}~\citep{xia2023moderate}~(abbreviated as Moderate). (vi). \textit{CCS}~\citep{zheng2023coveragecentric}. (vii). \textit{Probabilistic coreset}~\citep{zhou2022probabilistic}~(abbreviated as Probabilistic). Note that we do not compare our LBCS with the method~\citep{borsos2020coresets} that also works in bilevel coreset selection, since it suffers from huge time consumption~\citep{zhou2022probabilistic}. For every newly added example, the consumption increases rapidly with the coreset size. Also, as reported in~\citep{zhou2022probabilistic}, its performance is not very competitive compared with ``Probabilistic coreset''. Technical details of employed baselines are provided in Appendix~\ref{supp:baselines}. For fair comparisons, we reproduce the baselines based on their code repositories. All experiments are repeated ten times on NVIDIA GTX3090 GPUs with PyTorch.


\myPara{Datasets and implementation.} We employ Fashion-MNIST~(abbreviated as F-MNIST)~\citep{xiao2017fashion}, SVHN~\citep{netzer2011reading}, and CIFAR-10~\citep{krizhevsky2009learning} to evaluate our method. The three benchmarks are popularly used~\citep{killamsetty2021retrieve,yang2023dataset}. In the procedure of coreset selection, we employ a LeNet for F-MNIST, and simple convolutional neural networks~(CNNs) for SVHN and CIFAR-10. An Adam optimizer~\citep{kingma2015adam} is used with a learning rate of 0.001 for the inner loop. The parameters $\epsilon$ and $T$ are set to 0.2 and 500. After coreset selection, for training on the constructed coreset, we utilize a LeNet~\citep{lecun1998gradient} for F-MNIST, a CNN for SVHN, and a ResNet-18 network for CIFAR-10 respectively. In addition, for F-MNIST and SVHN, an Adam optimizer~\citep{kingma2015adam} is used with a learning rate of 0.001 and 100 epochs. For CIFAR-10, an SGD optimizer is exploited with an initial learning rate of 0.1 and a cosine rate scheduler. 200 epochs are set totally. Details of network architectures are given in Appendix~\ref{supp:networks}. 

\myPara{Measurements.} We consider two kinds of comparisons with the above baselines. (i). The same predefined coreset size is applied in the beginning. After coreset selection and model training on the constructed coreset,  measurements are both the model accuracy on test data and coreset size. A higher accuracy and smaller coreset size indicate better coreset selection. Comparing different methods of coreset selection, the average accuracy brought by per data point within the coreset is also provided. (ii). We apply the coreset size obtained by our method to the baselines. Their coreset selection and model training are then based on this coreset size. Measurements are the model accuracy on test data under the same coreset size. Here a higher accuracy means superior coreset selection.

\begin{table*}[!t]
    \centering
    \scriptsize
    \setlength{\tabcolsep}{2.5mm}{
    \begin{tabular}{l|c|ccccccc|>{\columncolor{Gray}}c}
    \toprule
     & $k$ &Uniform & EL2N & GraNd &Influential & Moderate & CCS & Probabilistic & LBCS~(ours) \\\midrule
    \parbox[t]{2mm}{\multirow{4}{*}{\rotatebox{90}{\tiny F-MNIST}}}& 956& 76.5$\pm$1.8 & 71.3$\pm$3.1 & 70.8$\pm$1.1  & 78.2$\pm$0.9 & 76.3$\pm$0.5 & 75.4$\pm$1.1 &79.2$\pm$0.9 & \textbf{79.7$\pm$0.5}\\
    &1935& 79.8$\pm$2.1 & 73.2$\pm$1.3 & 71.2$\pm$1.5 & 80.0$\pm$1.9 & 79.7$\pm$0.5 & 80.3$\pm$0.6 &81.7$\pm$0.7 & \textbf{82.8$\pm$0.4}\\
    &2832& 81.2$\pm$1.3 & 75.0$\pm$1.6 & 73.2$\pm$1.1 & 81.0$\pm$0.7 & 81.4$\pm$0.3 & 82.5$\pm$0.7 & 83.4$\pm$0.6 & \textbf{84.0$\pm$0.4}\\
    &3746& 82.8$\pm$1.5 & 77.0$\pm$2.2 & 75.1$\pm$1.6 & 82.1$\pm$1.0 & 82.2$\pm$0.4 & 83.6$\pm$1.0 & 83.8$\pm$0.5& \textbf{84.5$\pm$0.3}\\\midrule
    \parbox[t]{2mm}{\multirow{4}{*}{\rotatebox{90}{\tiny SVHN}}}& 970 & 66.7$\pm$2.6 & 57.2$\pm$0.5 & 60.6$\pm$1.7 & 70.3$\pm$1.2 & 68.4$\pm$1.8 & 65.1$\pm$1.1 &67.6$\pm$1.3 & \textbf{70.6$\pm$0.3}\\
    & 1902& 75.7$\pm$1.8 & 65.0$\pm$0.7 & 67.0$\pm$1.2 & 75.5$\pm$0.9 & 77.7$\pm$1.2 & 75.9$\pm$1.4 &76.1$\pm$0.7 & \textbf{78.3$\pm$0.7}\\
    & 2713& 79.5$\pm$2.6 & 72.3$\pm$0.5 & 74.8$\pm$1.1 & 80.0$\pm$1.9 & 81.4$\pm$1.1 & 81.1$\pm$1.0 & 80.5$\pm$0.4& \textbf{82.3$\pm$0.8}\\
    & 3805& 83.6$\pm$1.2 & 75.5$\pm$1.8 & 78.2$\pm$1.3 & 82.8$\pm$1.6 & 83.6$\pm$0.6 & 84.2$\pm$0.3 & 83.5$\pm$1.2 & \textbf{84.6$\pm$0.6}\\\midrule
    \parbox[t]{2mm}{\multirow{4}{*}{\rotatebox{90}{\tiny CIFAR-10}}}& 970& 46.8$\pm$1.2 & 36.7$\pm$1.1 & 41.4$\pm$1.9 & 44.8$\pm$1.5 &46.2$\pm$1.9 &45.4$\pm$1.0 &  47.8$\pm$1.1 & \textbf{48.3$\pm$1.2}\\
    & 1955 & 58.0$\pm$1.3 & 48.3$\pm$1.9 & 52.5$\pm$1.2 & 57.6$\pm$1.9 & 57.4$\pm$0.8 & 58.6$\pm$1.4 & 59.4$\pm$1.2 & \textbf{60.4$\pm$1.0}\\
    & 2914 & 65.5$\pm$1.9 & 55.0$\pm$3.2 & 67.7$\pm$1.8 & 67.2$\pm$1.0 & 68.2$\pm$2.1 & 66.5$\pm$1.0 & 68.0$\pm$0.8 & \textbf{69.5$\pm$0.9}\\
    & 3736 & 70.6$\pm$2.4 & 58.8$\pm$1.9 & 72.8$\pm$1.1 & 70.2$\pm$3.5 & 73.0$\pm$1.2 & 72.8$\pm$0.9  &\textbf{73.4$\pm$0.5} & \textbf{73.4$\pm$0.5}\\
    \bottomrule
    \end{tabular}}
    \caption{Mean and standard deviation of test accuracy (\%) on different benchmarks with \textit{coreset sizes achieved by the proposed LBCS}.}
    \label{tab:exact_results_1}
\end{table*}

\myPara{Discussions on experimental results.}
Results about the first kind of comparison are provided in Table~\ref{tab:exact_results}. As can be seen, for SVHN, our method always achieves the best test accuracy meanwhile with smaller coreset sizes compared with predefined ones. For F-MNIST and CIFAR-10, our LBCS obtains the best accuracy with the smaller coreset sizes most of the time. When $k=1000$ on F-MNIST and $k=4000$ on CIFAR-10, our performance is competitive (80.3$\pm$0.6 vs. 79.7$\pm$0.5 and 73.9$\pm$0.4 vs. 73.4$\pm$0.5). Also, based on the results of the first kind of comparison, we provide the average accuracy brought by per data point within the coreset in Appendix~\ref{supp:mean_per_data_point}, which shows that our LBCS always enjoys higher average accuracy. In addition, results of the second kind of comparison are provided in Table~\ref{tab:exact_results_1}. Clearly, our LBCS consistently outperforms all competitors. Based on these observations, we can safely conclude that our method can reach competitive model performance with smaller coreset sizes, or better model performance with the same coreset sizes.

\subsection{Robustness against Imperfect Supervision}\label{sec:imperfect}

\begin{figure}[!t]
    \centering
    \subfigure[]{
    \begin{minipage}[b]{0.21\textwidth}
			\includegraphics[width=1.05\textwidth]{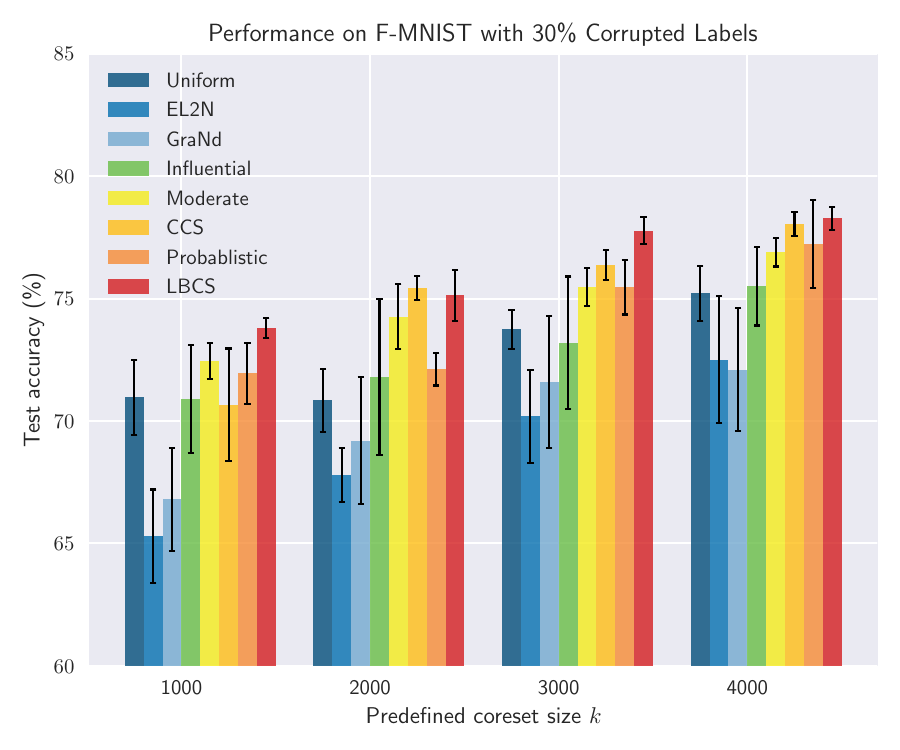} 
	\end{minipage}
  \label{fig:im_a}}
 \subfigure[]{
 \begin{minipage}[b]{0.21\textwidth}
			\includegraphics[width=1.05\textwidth]{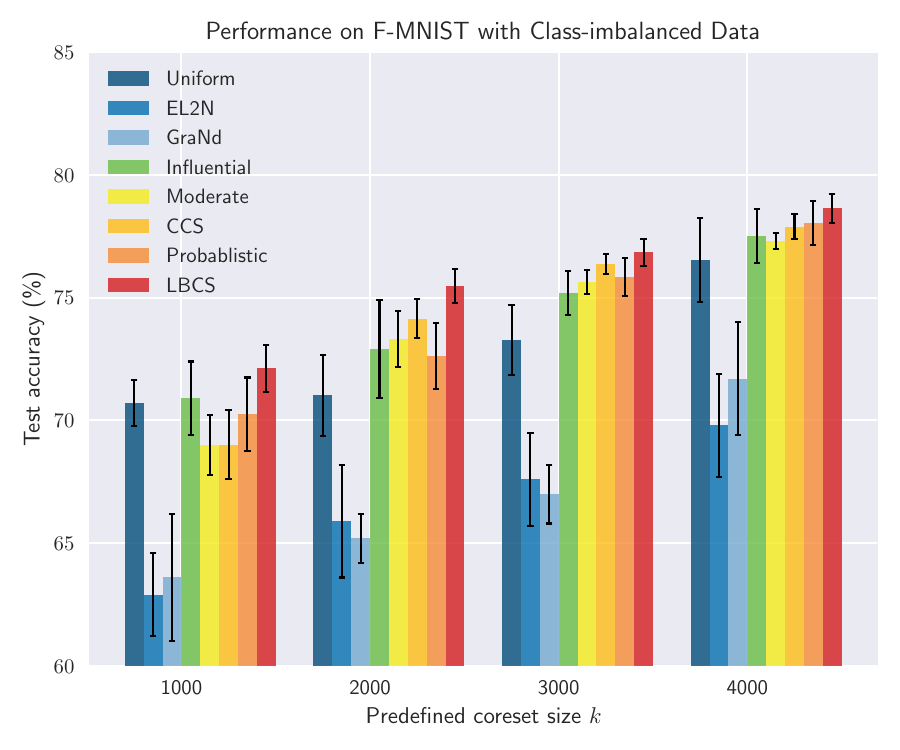} 
	\end{minipage}
 \label{fig:im_c}}
 \vspace{-6pt}
    \caption{Illustrations of coreset selection under imperfect supervision. \textbf{(a)} Test accuracy~(\%) in coreset selection with 30\% corrupted labels; \textbf{(b)} Test accuracy~(\%) in coreset selection with class-imbalanced data. The optimized coreset sizes by LBCS in these cases are provided in Appendix~\ref{supp:im_coreset_size}.} 
    \vspace{-8pt}
\end{figure}

\myPara{Coreset selection with corrupted labels.} We employ F-MNIST here. We inject 30\% symmetric label noise~\citep{ma2020normalized,kim2021fine,park2023robust} into the original clean F-MNIST to generate the noisy version of F-MNIST. Namely, the labels of 30\% training data are flipped. The predefined coreset size $k$ is set to 1000, 2000, 3000, and 4000 respectively. Experimental results are provided in Figures~\ref{fig:im_a}. The results support our claims made in \textbf{Remark~2}, which demonstrate that LBCS can reduce the model overfitting in coreset selection and help model generalization. We also evaluate LBCS when the noise level is higher, \ie, 50\%. Results can be found in Appendix~\ref{supp:50_noise}.

\myPara{Coreset selection with class-imbalanced data.} For the class-imbalanced experiment, we adopt a similar setting as in~\citep{xu2021towards}. The exponential type of class imbalance~\citep{cao2019learning} is used. The imbalanced ratio is set to 0.01. Experimental results are provided in Figure~\ref{fig:im_c}, which confirms the validity of our method in coreset selection with class-imbalanced cases.

\subsection{Evaluations on ImageNet-1k}

We evaluate the performance of LBCS  on ImageNet-1k~\citep{deng2009imagenet}. The network structures for the inner loop and training on the coreset after coreset selection are ResNet-50. As the size of ImageNet-1k is huge, to accelerate coreset selection, the tricks discussed previously are employed. We regard 100 examples as a group, where they share the same mask. The same tricks are applied to the baseline Probabilistic. Staying with precedent~\citep{sorscher2022beyond}, the VISSL library~\citep{goyal2021vissl} is used. Besides, for model training after coreset selection, we adopt a base learning rate of 0.01, a batch size of 256, an SGD optimizer with a momentum of 0.9, and a weight decay of 0.001. 100 epochs are set in total. The experiment in each case is performed once, considering calculation costs. We set the predefined \textit{ratio} of coreset selection, \ie, $k/n$, to 70\% and 80\% respectively. Experimental results are provided in Table~\ref{tab:imagenet}, which confirm the validity of our LBCS.

\begin{table}[!t]
    \centering
    \footnotesize
    \begin{tabular}{l|cc}
    \toprule
        $k/n$ & 70\% & 80\% \\
        \midrule
        Uniform & 88.63 & 89.52\\
        EL2N & 89.82 & 90.34\\
        GraNd & 89.30 & 89.94\\
        Influential & - & -\\
        Moderate & 89.94 & 90.65\\
        CCS & 89.45 & 90.51\\
        Probabilistic & 88.20 & 89.35\\\midrule
        \rowcolor{Gray}LBCS (ours) & \textbf{89.98~(68.53\%)} & \textbf{90.84~(77.82\%)}\\
        \bottomrule
    \end{tabular}
    \caption{Top-5 test accuracy (\%) on ImageNet-1k. Partial results are from previous work~\citep{xia2023moderate}. The best test accuracy in each case is in \textbf{bold}. For LBCS, we additionally report the optimized ratio of coreset selection.}
    \vspace{-8pt}
    \label{tab:imagenet}
\end{table}

\section{More Justifications and Analyses}

\myPara{The influence of the number of search times.} Here we investigate the number of search times during coreset selection, \ie, the value of $T$. We conduct experiments on F-MNIST. Experimental results are provided in Appendix~\ref{supp:search_time}. The main observation is that, at the beginning, with the increased search times, the test accuracy increases, and the coreset size decreases. As the search proceeds, the test accuracy gradually stabilizes. The coreset size continues to be smaller. Afterward, when the number of searches is large, the search results are not changed obviously, as the search approaches convergence empirically. In practice, we can pick a suitable $T$ based on the need for coresets and the budget of search in coreset selection. 

\myPara{Time complexity analysis.} We compare the time complexity between Probabilistic~\citep{zhou2022probabilistic} and our LBCS, because both the two methods are based on the bilevel framework for coreset selection. Specifically, suppose that the number of training epochs of one inner loop is denoted as $K$. The time complexity of our LBCS is $\mathcal{O}(TK)$. As a comparison, the time complexity of Probabilistic~\citep{zhou2022probabilistic} is $\mathcal{O}(TKC)$, where $C$ is the number of sampling times required by its policy gradient estimator. As the value of $C$ is generally greater than 1~\citep{zhou2022probabilistic}, our LBCS enjoys less time complexity. 

In addition to the above analysis, we also explore the superiority of our method in the case of \textit{cross network architectures}. That is to say, the architectures used for coreset selection on training data  and evaluations on test data are different. We employ ViT~\citep{dosovitskiy2021image} and WideResNet~\citep{zagoruyko2016wide} (see Appendix~\ref{supp:network_results}). Moreover, the evaluations about two applications of coreset selection, \ie, \textit{continual learning}~\citep{kim2022a} and \textit{streaming}~\citep{hayes2019memory}, can be found in Appendix~\ref{supp:cl_results} and Appendix~\ref{supp:streaming_results} respectively.

        \vspace{-7pt}
\section{Conclusion}\label{sec:conclusion}
\vspace{-2pt}
In this paper, we propose and delve into the problem of refined coreset selection. An advanced method named lexicographic bilevel coreset selection is presented. We theoretically prove its convergence and conduct comprehensive experiments to demonstrate its superiority. For future work, we are interested in adapting the proposed method to other fields such as image and motion generation~\citep{song2023consistency,chen2023executing}, and in accelerating the pre-training of large vision and language models~\citep{touvron2023llama,liu2023visual} with our method. 

\section{Impact Statement}
This paper presents work on the problem of refined coreset selection~(RCS), which is significant in this big data era. A framework of lexicographic bilevel coreset selection is proposed for the problem, with both theoretical guarantees and superior performance. The outcome of this paper has several broader impacts as follows. For example, due to data privacy and security, data sharing is often challenging. With the outcome of the coreset selection by this paper, data sharing can be promoted by only sharing representative data in the constructed coreset, but not full data. Besides, the outcome of this paper helps sustainable development, since it can lower the energy and physical resource requirements of machine learning algorithms, which reduces their impact on the environment. The RCS problem is realistic and important. The solution for it is non-trivial. Therefore, the development and realization of the algorithm for RCS require advanced technology and expertise, which may result in the emergence of technical barriers.

	
	\bibliography{draft}
	\bibliographystyle{icml2024}

	\clearpage
\onecolumn
\appendix

\etocdepthtag.toc{mtappendix}
\etocsettagdepth{mtchapter}{none}
\etocsettagdepth{mtappendix}{subsection}

\renewcommand{\contentsname}{Appendix}
\tableofcontents

\clearpage

\section{Details of the Black-box Optimization Algorithm}
\label{supp:lexiflow}
\myPara{Technical details.} For the black-box optimization of $f_1$ and $f_2$ in order of priority, we make use of a randomized direct search algorithm named LexiFlow~\citep{zhang2023targeted} and make necessary modifications to it\footnote{We remove optional
input targets and adjust compromise from an absolute value to a relative value.}. In RCS, LexiFlow is designed to iteratively direct the search to the optimal solution based on lexicographic comparisons over pairs of masks. Technically, at the $i$-th iteration, LexiFlow maintains an incumbent point that represents the optimal mask up to the $i$-th iteration. The algorithm will sample two new masks near the incumbent point and update the incumbent point by making lexicographic comparisons between the incumbent point and sampled masks. During the iterative optimization process, LexiFlow will gradually move toward the optimal solution.
To free the algorithm from local optima and manual configuration of the step size, LexiFlow includes restart and dynamic step size techniques. These techniques are similar to those used in an existing randomized direct search-based method~\citep{wu2021frugal}. The details are provided in Algorithm~\ref{alg:lexiflow}. 

\begin{algorithm}[H]\small
\SetNoFillComment
\SetAlCapNameFnt{\small}
\SetAlCapFnt{\small}
\caption{Lexicographic Optimization for $f_1$ and $f_2$.}
\label{alg:lexiflow}
\KwIn{
Objectives $F(\cdot)$, compromise $\epsilon$.
}
\textbf{Initialization:} Initial mask $\bm{m}_0$, $t' = r= e = 0$, and $\delta = \delta_{\rm{init}}$; \\
$\bm{m}^* \gets \bm{m}_0 $, $\cH \gets \{ \bm{m}_0 \}$, and $F_{\cH} \gets F(\bm{m}_0)$. \\
\While{$t = 0, 1, ...$}{
Sample $\bm{u}$ uniformly from unit sphere $\mathbb{S}$;

\lIf{\rm{update}~($F(\bm{m}_t + \delta \bm{u})$,  $F{(\bm{m}_{t})}$, $F_{\cH}$)}{$\bm{m}_{t+1} \gets \bm{m}_{t} + \delta \bm{u}$, $t' \gets t$}
\lElseIf{\rm{update}~($F(\bm{m}_{t}- \delta\bm{u})$,  $F{(\bm{m}_{t})}$, $F_{\cH}$)}{
    $\bm{m}_{t+1} \gets \bm{m}_t - \delta \bm{u}$, $t' \gets t$}
\lElse{
    $\bm{m}_{t+1} \gets \bm{m}_t$, 
    $e \gets e +1$}
$\mathcal{H} \gets \mathcal{H} \cup \{\bm{m}_{t+1}\}$, and update $F_{\cH}$ according to \eqref{eq:lexi-target-input}
\lIf{$e = {2^{n-1}}$}
{$e \gets 0, 
\delta \gets \delta \sqrt{(t'+1)/(t+1)}$} 
\If{$\delta < \delta_{\rm lower}$}{ \tcp{Random restart;} 
$r \gets r+1$, $\bm{m}_{t+1} \gets \mathcal{N}(\bm{m}_0, \bm{I})$, $\delta \gets \delta_{\rm{init}} + r$;}
}
~~\textbf{Procedure} \rm{update}~($F(\bm{m}')$,  $F{(\bm{m})}$, $F_{\cH}$): \\
~~~~~~~~~~\If{$F(\bm{m}') \vec{\prec}_{(F_{\cH})} F(\bm{m})$ \rm{\textbf{or}} \big($F(\bm{m}') \vec{=}_{(F_{\cH}) } F(\bm{m})$ and $F(\bm{m}') \vec{\prec} F(\bm{m})\big)$}{
~~~~~~~\If{$F(\bm{m}') \vec{\prec}_{(F_{\cH})} F({\bm{m}^*})$ \rm{\textbf{or}} \big($F(\bm{m}') \vec{=}_{(F_{\cH}) } F({\bm{m}^*})$ and $F(\bm{m}') \vec{\prec}_l F({\bm{m}^*})$\big)}{~~~~~~~~$\bm{m}^* \gets \bm{m}'$; ~~}
~~~~~~~\textbf{Return True} ~~~~~~~~~~~~~~~~~~~~}
~~\Else{\textbf{Return False} ~~~~~~~~~~~~~~~} 
\textbf{Output:} The optimal mask $\bm{m}^*$.
\end{algorithm}

\myPara{Practical lexicographic relations.} We highlight that the notations of lexicographic relations in Algorithm~\ref{alg:lexiflow}~(\ie, $\vec{=}_{(F_{\cH})}$, $\vec{\prec}_{(F_{\cH})}$, and  $\vec{\preceq}_{(F_{\cH})}$) are a bit different from those in the main paper. It is because the optimization with the lexicographic relations in Definition~\ref{def:lexico_relations} relies on the infimums of $f_1(\bm{m})$ and $f_2(\bm{m})$. They are theoretically achievable but may be inaccessible in practice. Therefore, in experiments, we use practical lexicographic relations that are defined with the available minimum values of objectives.

Specifically, given any two masks $\bm{m}'$ and $\bm{m}$, the practical lexicographic relations $\vec{=}_{(F_{\cH})}$, $\vec{\prec}_{(F_{\cH})}$, and  $\vec{\preceq}_{(F_{\cH})}$ in Algorithm~\ref{alg:lexiflow} are defined as:
\begin{align}  
 & F(\bm{m})  \vec{=}_{(F_{\cH})} F(\bm{m}')  \Leftrightarrow \forall i \in [2]:   f_{i}(\bm{m}) =  f_{i}(\bm{m}')  \lor  
 (f_{i}(\bm{m})   \leq \tilde{f}_i^* \land f_{i}(\bm{m}') \leq  \tilde{f}_i^* ), \\
 &  F(\bm{m})  \vec{\prec}_{(F_{\cH})}  F(\bm{m}')   \Leftrightarrow  
  \exists i \in [2]: f_{i}(\bm{m}) < f_{i}(\bm{m}')  \land  f_{i}(\bm{m}') > \tilde{f}_i^*  \land F_{i-1}(\bm{m}) \vec{=}_{(F_{\cH})} F_{i-1}(\bm{m}'),
\\ 
 &  F(\bm{m}) \vec{\preceq}_{(F_{\cH})} F(\bm{m}')   \Leftrightarrow  F(\bm{m}) \vec{\prec}_{(F_{\cH})} F(\bm{m}') \lor  F(\bm{m}) \vec{=}_{(F_{\cH})} F(\bm{m}'),
\end{align}
where $F_{i-1}(\bm{m})$ denotes the a vector with the first $i-1$ dimensions of $F(\bm{m})$, \ie, $F_{i-1}(\bm{m}) = [f_{1}(\bm{m}),...,f_{i-1}(\bm{m})]$. 
The optimizing thresholds for each objective are represented by $F_{\cH} = [\tilde{f}_1^*,\tilde{f}_2^*]$, signifying that any masks achieving these thresholds can be considered equivalent with respect to the given objective.
That $\tilde{f}_i^*$ is computed based on historically evaluated points $\mathcal{H}$. 
Given $\mathcal{M}_{\mathcal{H}}^{0} = \mathcal{H}$, we further have:
\begin{align}\label{eq:lexi-target-input}
    &\mathcal{M}_{\mathcal{H}}^{1}:=\{\bm{m}\in \mathcal{M}_{\mathcal{H}}^{0}~|~f_1(\bm{m})\leq \tilde{f}_1^*\},\ \ \hat{f}_1^*:=\inf_{\bm{m}\in \mathcal{M}_{\mathcal{H}}^{0} }f_1(\bm{m}),\ \ \tilde{f}_1^* = \hat{f}_1^**(1+\epsilon), \\
    &\mathcal{M}_{\mathcal{H}}^{2}:=\{\bm{m}\in \mathcal{M}_{\mathcal{H}}^{1}~|~f_2(\bm{m})\leq \tilde{f}_2^*\},  \ \ \hat{f}_2^*:=\inf_{\bm{m}\in \mathcal{M}_{\mathcal{H}}^{1}}f_2(\bm{m}),\ \ \text{and} \ \ \tilde{f}_2^* =\hat{f}_2^*.\nonumber
\end{align}

\section{Proofs of Theoretical Results}
\label{proof:convergence}

The proof of Theorem~1 is detailed as follows.

\begin{proof} 
We use $\bm{m}^0$ to denote the mask generated at the step $0$, where the mask $\bm{m}^0 \notin \mathcal{M}_{1}^{*}$ and $\bm{m}^0 \notin \mathcal{M}_{2}^{*}$. We use $d_{f_{i}}(\bm{a},\bm{b})$ to denote the difference between the mask $\bm{a}$ and the mask $\bm{b}$ on the optimization objective $f_{i}$, \ie,
\begin{align}
    d_{f_{i}}(\bm{a},\bm{b}) = |f_{i}(\bm{a})-f_{i}(\bm{b})| ~~\forall \bm{a},\bm{b} \in \mathcal{M}.
\end{align}
Given Condition~\ref{condition:lower-bound}, there exists $n_{1} \in \mathbb{R}^{+}$, $n_{2} \in \mathbb{R}^{+}$ for $f_{1}$ and $f_{2}$ such that:
\begin{align}
\label{eq:max_dis}
    &d_{f_{1}}(\bm{a},\bm{b}) < n_{1}\gamma_{1} ~~~\forall \bm{a},\bm{b} \in \mathcal{S}_1, \\ 
\label{eq:max_dis2}    
    &d_{f_{2}}(\bm{a},\bm{b}) < n_{2}\gamma_{2} ~~~\forall \bm{a},\bm{b} \in \mathcal{S}_2,
\end{align}
in which $\mathcal{S}_{1} = \{ \bm{m}^{t} \in \mathcal{M} | f(\bm{m}^{t}) \leq f(\bm{m}^{0}) \}$ and $\mathcal{S}_{2} = \{ \bm{m}^{t} \in \mathcal{M} | f(\bm{m}^{t}) \leq f(\bm{m}^{\hat{t}}) \}$ as stated in Condition~\ref{condition:lower-bound}. Intuitively speaking, (\ref{eq:max_dis}) and (\ref{eq:max_dis2}) imply that it needs at most $n_{1}$ and $n_{2}$ time steps for the mask $\bm{a}$ to surpass the mask $\bm{b}$ in optimization objectives $f_{1}$ and $f_{2}$, respectively.


LBCS consists of two types of optimization stages, including a stage where the first objective $f_{1}$ is optimized, and a stage where the second objective $f_{2}$ is optimized while ensuring that $f_{1}$ remains within the optimal region with the compromise $\epsilon$. We thus analyze the convergence behavior of LBCS according to these two stages.  

\myPara{$f_{1}$ optimization stage:}

We define $\bm{m}_{1}^{*}:= \mathop{\arg\max}\limits_{\bm{m} \in \mathcal{M}_{1}^{*}}\{f_{1}(\bm{m})\}$. By substituting $\bm{m}^0$ and $\bm{m}_{1}^{*}$ into $\bm{a}$ and $\bm{b}$ in Eq.~(\ref{eq:max_dis}), we have:
\begin{align} 
\label{eq:dis_tobest}
    d_{f_{1}}(\bm{m}^{0},\bm{m}_{1}^{*}) < n_{1}\gamma_{1}. 
\end{align}
According to Condition~\ref{condition:lower-bound}, we have \(n_1 \in \mathbb{R}^+ \)  and $0< \eta_{1} \leq 1$ such that,
\begin{align}
    \mathbbm{P}(f_{1}(\bm{m}^{n_{1}}) \leq f_{1}(\bm{m}_{1}^{*})) &= \mathbbm{P}(f_{1}(\bm{m}^{n_{1}}) \leq f_{1}^* * (1 + \epsilon)) \\ \nonumber
    &= \mathbbm{P}(\bm{m}^{n_{1}} \in \mathcal{M}_1^*) \\ \nonumber
    &\geq \eta_{1}^{n_{1}}.
\end{align}
For $j = 1,2,...$, we have:
\begin{align}
\label{eq:n1_1}
\mathbbm{P}(f_{1}(\bm{m}^{jn_{1}}) \leq  f_{1}(\bm{m}_{1}^{*}) ) &= \mathbbm{P}(\bm{m}^{jn_1} \in \mathcal{M}_1^*) \\ \nonumber
&=1- \mathbbm{P}(\bm{m}^{jn_1} \notin \mathcal{M}_1^*) \\ \nonumber
&\geq 1-(1-\eta_{1}^{n_{1}})^j. \nonumber
\end{align}
According to Condition~\ref{condition:progressable}, $\bm{m}^{1}, ..., \bm{m}^{n_{1}-1}$ all belong to $\mathcal{S}_1$, $\forall i \in [n_1-1]$,
\begin{align}
\label{eq:n1_2}
\mathbbm{P}(f_{1}(\bm{m}^{jn_{1}+i}) \leq  f_{1}(\bm{m}_{1}^{*}) )&=\mathbbm{P}(\bm{m}^{jn_1+i} \in \mathcal{M}_1^*)  \\ \nonumber
& = 1- \mathbbm{P}(\bm{m}^{jn_1+i} \notin \mathcal{M}_1^*) \\ \nonumber
&\geq 1-(1-\eta_{1}^{n_1})^j.
\end{align}
When $j$ tends to $+\infty$, $1-(1-\eta_{1}^{n_{1}})^j$ tends to $1$. Then, combining (\ref{eq:n1_1}) and (\ref{eq:n1_2}), the algorithm will reach $\mathcal{M}_1^*$. 

\myPara{The optimization of $f_{2}$ in the set $\mathcal{M}_{1}^*$:}

We use $\hat{t}$ to denote the time step that the algorithm reaches $\mathcal{M}_1^*$. 
Also, we define $\bm{m}_{2}^{*} := \mathop{\arg\max}_{\bm{m} \in \mathcal{M}_{2}^{*}}\{f_{2}(\bm{m})\}$. By substituting $\bm{m}^{\hat{t}}$ and $\bm{m}_{2}^{*}$ into $\bm{a}$ and $\bm{b}$ in (\ref{eq:max_dis2}), we have:
\begin{align} 
\label{eq:dis_tobest2}
    d_{f_{2}}(\bm{m}^{\hat{t}},\bm{m}_{2}^{*}) < n_{2}\gamma_{2}.
\end{align}
According to Condition~\ref{condition:lower-bound}, we have  \(n_2 \in \mathbb{R}^+ \), and $0 < \eta_{2} \leq 1$ such that: 
\begin{align}
\mathbbm{P}(\bm{m}^{\hat{t}+n_{2}} \in \mathcal{M}_2^*) \geq \eta_{2}^{n_{2}},
\end{align}
while the mask sequence $\{\bm{m}^t\}_{t=\hat{t}+1}^{t=\hat{t}+n_{2}}$ satisfies:
\begin{align}
\forall \bm{m} \in \{\bm{m}^t\}_{t=\hat{t}+1}^{t=\hat{t}+n_{2}}: f_1(\bm{m}) \in \mathcal{M}_{1}^*.
\end{align}
For $j = 1,2,...$, we have:
\begin{align}
\label{eq:n2_1}
\mathbbm{P}(f_2(\bm{m}^{j(\hat{t}+n_{2})}) \leq  f_{2}(\bm{m}_{2}^{*}) ) &= \mathbbm{P}(f_2(\bm{m}^{j(\hat{t}+n_{2})}) \leq f_{2}^*) \\ \nonumber
&= \mathbbm{P}(\bm{m}^{j(\hat{t}+n_{2})} \in \mathcal{M}_2^*)  \\ \nonumber
&= 1-\mathbbm{P}(\bm{m}^{j(\hat{t}+n_{2})} \notin \mathcal{M}_2^*) \\ \nonumber
& \geq 1-(1-\eta_{2}^{n_{2}})^j.
\end{align}
According to Condition~\ref{condition:progressable}, $\bm{m}^{\hat{t}+1}, ..., \bm{m}^{\hat{t}+n_{2}-1}$ all belongs to $\mathcal{S}_{2}$, $\forall i \in [n_{2}-1]$,
\begin{align}
\label{eq:n2_2}
\mathbbm{P}(f_2(\bm{m}^{j(\hat{t}+n_{2})+i}) \leq  f_{2}(\bm{m}_{2}^{*}) )  &= \mathbbm{P}(f_2(\bm{m}^{j(\hat{t}+n_{2})+i}) \leq f_{2}^*) \\ \nonumber
&=  \mathbbm{P}(\bm{m}^{j(\hat{t}+n_{2})+i} \in \mathcal{M}_2^*) \\ \nonumber
& = 1- \mathbbm{P}(\bm{m}^{j(\hat{t}+n_{2})+i} \notin \mathcal{M}_2^*) \\  \nonumber
& \geq 1-(1-\eta_{2}^{n_2})^j.
\end{align}
When $j$ tends to $+\infty$, $1-(1-\eta_{2}^{n_2})^j$ tends to 1. Afterward, combining (\ref{eq:n2_1}) and (\ref{eq:n2_2}), the algorithm will reach $\mathcal{M}_2^*$. 
Proof complete.

\end{proof}

\section{Supplementary Notes of Probabilistic Bilevel Coreset Selection}\label{supp:gradients}
\subsection{Method Description}
Previous work~\citep{zhou2022probabilistic} proposes probabilistic bilevel coreset selection, which continualizes weights by probabilistic reparameterization for ease of optimization. Specifically, the mask $m_i$ is reparameterized as a Bernoulli random variable with probability $s_i$ to be 1 and $1-s_i$ to be 0. Namely, $m_i\sim\text{Bern}(s_i)$, where $s_i\in[0,1]$. If we assume that the variables $m_i$ are independent, the distribution function of $\bm{m}$ can be denoted as $p(\bm{m}|\bm{s})=\prod_{i=1}^n(s_i)^{m_i}(1-s_i)^{(1-m_i)}$. Besides, the coreset size can be controlled by the sum of the probabilities $s_i$, as $\mathbbm{E}_{\bm{m}\sim p(\bm{m}|\bm{s})}\|\bm{m}\|_0=\sum_{i=1}^n s_i=\bm{1}^\top\bm{s}$. Afterward, combining the definition of $f_1(\bm{m})$, the original probabilistic bilevel coreset selection~\citep{zhou2022probabilistic} can be formulated as 
\begin{align}
    \min_{\bm{s}} \mathbbm{E}_{p(\bm{m}|\bm{s})}f_1(\bm{m}), \ \ \text{s.t.}\ \ \bm{\theta}(\bm{m})\in\arg\min_{\bm{\theta}}\mathcal{L}(\bm{m},\bm{\theta}). 
\end{align}
By introducing $f_2(\bm{m})$, the probabilistic bilevel coreset selection is modified to 
\begin{align}\label{eq:pro_bi_m}
    \min_{\bm{s}} \mathbbm{E}_{p(\bm{m}|\bm{s})}f_1(\bm{m})+\mathbbm{E}_{p(\bm{m}|\bm{s})}f_2(\bm{m}), \ \ \text{s.t.}\ \ \bm{\theta}(\bm{m})\in\arg\min_{\bm{\theta}}\mathcal{L}(\bm{m},\bm{\theta}). 
\end{align}
\subsection{Gradient Analysis}
We derive the gradients of the outer loop of (\ref{eq:cs_bi_mpo}) as 
\begin{align}\label{eq:gradient_pro_bi_m}
\nabla_{\bm{s}}\left[\mathbbm{E}_{p(\bm{m}|\bm{s})}f_1(\bm{m})+\mathbbm{E}_{p(\bm{m}|\bm{s})}f_2(\bm{m})\right]&=\nabla_{\bm{s}}\int f_1(\bm{m})p(\bm{m}|\bm{s})\mathrm{d}\bm{m} + \nabla_{\bm{s}}\mathbbm{E}_{p(\bm{m}|\bm{s})}\|\bm{m}\|_0\\\nonumber
&=\int f_1(\bm{m})\frac{\nabla_{\bm{s}}p(\bm{m}|\bm{s})}{p(\bm{m}|\bm{s})}p(\bm{m}|\bm{s})\mathrm{d}\bm{m}+\nabla_{\bm{s}}\mathbf{1}^\top\bm{s}\\\nonumber
&=\int f_1(\bm{m})\nabla_{\bm{s}}\ln p(\bm{m}|\bm{s})p(\bm{m}|\bm{s})\mathrm{d}\bm{m} + \nabla_{\bm{s}}\mathbf{1}^\top\bm{s}\\\nonumber
&=\mathbbm{E}_{p(\bm{m}|\bm{s})}f_1(\bm{m})\nabla_{\bm{s}}\ln p(\bm{m}|\bm{s})+\mathbf{1}.\\\nonumber
\end{align}
In the last line of (\ref{eq:gradient_pro_bi_m}), the first term denotes the gradient of $\mathbbm{E}_{p(\bm{m}|\bm{s})}f_1(\bm{m})$ and the second term denotes the gradient of $\mathbbm{E}_{p(\bm{m}|\bm{s})}f_2(\bm{m})$. In optimization, we directly employ $f_1(\bm{m})\nabla_{\bm{s}}\ln p(\bm{m}|\bm{s})$, since it is an unbiased stochastic gradient of $\nabla_{\bm{s}}\mathbbm{E}_{p(\bm{m}|\bm{s})}f_1(\bm{m})$~\citep{zhou2022probabilistic}. We further derive that 
\begin{align}
    f_1(\bm{m})\nabla_{\bm{s}}\ln p(\bm{m}|\bm{s})&=f_1(\bm{m})\cdot\left(\frac{\bm{m}}{\bm{s}}-\frac{1-\bm{m}}{1-\bm{s}}\right)\\\nonumber
    &=f_1(\bm{m})\cdot\frac{\bm{m}(1-\bm{s})-\bm{s}(1-\bm{m})}{\bm{s}(1-\bm{s})}\\\nonumber
    &=f_1(\bm{m})\cdot\frac{\bm{m}-\bm{s}}{\bm{s}(1-\bm{s})}.
\end{align}
The gradient norms of two terms hence are $\|f_1(\bm{m})\cdot\frac{\bm{m}-\bm{s}}{\bm{s}(1-\bm{s})}\|_2$ and $\|\mathbf{1}\|_2=\sqrt{n}$ respectively. Therefore, the gradient forms of $(1-\lambda)\mathbbm{E}_{p(\bm{m}|\bm{s})}f_1(\bm{m})$ and $\lambda\mathbbm{E}_{p(\bm{m}|\bm{s})}f_2(\bm{m})$ are $(1-\lambda)\|f_1(\bm{m})\cdot\frac{\bm{m}-\bm{s}}{\bm{s}(1-\bm{s})}\|_2$ and $\lambda\|\mathbf{1}\|_2=\lambda \sqrt{n}$ respectively. In the following, for simplicity, we denote 
\begin{equation}
    \zeta_1(\lambda):=(1-\lambda)\|f_1(\bm{m})\cdot\frac{\bm{m}-\bm{s}}{\bm{s}(1-\bm{s})}\|_2~~\text{and}~~\zeta_2(\lambda):=\lambda \sqrt{n}.
\end{equation}
Clearly, the value of $\zeta_1(\lambda)$ depends on $f_1(\bm{m})$, $\bm{m}$, and $\bm{s}$, which is related to data, networks, optimization algorithms, and specific tasks simultaneously. The value of $\zeta_2(\lambda)$ is also related to data. This causes it to be hard to tune a suitable weight in practice. When $\lambda$ is set to $\frac{1}{2}$, $\zeta_2(\frac{1}{2})$ is large for the full optimization of $f_2$, since $\zeta_2(\frac{1}{2})=\frac{\sqrt{n}}{2}$ and $n$ is usually large in the task of coreset selection. Therefore, the coreset size will be minimized too much.

\subsection{Settings for Experiments in Figure~1}\label{supp:exp_fig1}
For the experiments in Figure~1, we employ a subset of MNIST. A convolutional neural network stacked with two blocks of convolution, dropout, max-pooling, and ReLU activation is used. Following~\citep{zhou2022probabilistic}, for the inner loop, the model is trained for
100 epochs using SGD with a learning rate of 0.1 and momentum of 0.9. For the outer loop, the probabilities are optimized
by Adam with a learning rate of 2.5 and a cosine scheduler.

\section{Supplementary Descriptions of Baselines and Network Structures}\label{supp:settings}

\subsection{Details of Baselines}\label{supp:baselines}
Below, we detail the used baselines in this paper.
\begin{itemize}[leftmargin=6mm]
    \item \textbf{Uniform sampling}. For this baseline, we randomly select partial data from full data to construct a coreset.
    \item \textbf{EL2N}~(NeurIPS 2021)~\citep{paul2021deep}\footnote{\href{https://github.com/mansheej/data_diet}{https://github.com/mansheej/data\_diet}}. The method involves the data points with larger norms of the error vector that is the predicted class probabilities minus one-hot label encoding.
    \item \textbf{GraNd}~(NeurIPS 2021)~\citep{paul2021deep}\footnote{\href{https://github.com/mansheej/data_diet}{https://github.com/mansheej/data\_diet}}. The method builds a coreset by involving the data points with larger loss gradient norms during training.
    \item \textbf{Influential coreset}~(ICLR 2023)~\citep{yang2023dataset}\footnote{\href{https://shuoyang-1998.github.io/assets/code/code_datasetptuning.zip}{https://shuoyang-1998.github.io/assets/code/code\_datasetptuning.zip}}. This algorithm utilizes the influence function~\citep{hampel1974influence}. The examples that yield strictly constrained generalization gaps are included in the coreset.
    \item \textbf{Moderate coreset}~(ICLR 2023)~\citep{xia2023moderate}\footnote{\href{https://github.com/tmllab/Moderate-DS}{https://github.com/tmllab/Moderate-DS}}. This method chooses the examples with the scores close to the score median in coreset selection. The score is about the distance of an example to its class center.
    \item \textbf{CCS}~(ICLR 2023)~\citep{zheng2023coveragecentric}\footnote{\href{https://github.com/haizhongzheng/Coverage-centric-coreset-selection}{https://github.com/haizhongzheng/Coverage-centric-coreset-selection}}. The method proposes a novel one-shot coreset selection method that jointly considers overall data coverage upon a distribution as well as the importance of each example. 
    \item \textbf{Probabilistic coreset}~(ICML 2022)~\citep{zhou2022probabilistic}\footnote{\href{https://github.com/x-zho14/Probabilistic-Bilevel-Coreset-Selection}{https://github.com/x-zho14/Probabilistic-Bilevel-Coreset-Selection}}. The method proposes continuous probabilistic bilevel optimization for coreset selection. A solver is developed for the bilevel optimization problem via unbiased policy gradient without the trouble of implicit differentiation.
\end{itemize}

\subsection{Details of Network Structures}\label{supp:networks}
We provide the detailed network structures of the used models in our main paper, which can be checked in Table~\ref{tab:network_structures}.

\begin{table}[!t]
    \centering
     \footnotesize
    \begin{tabular}{c|c|c}
    \hline
    CNN for SVHN~(inner loop) & CNN for SVHN (trained on coresets) & CNN for CIFAR-10~(inner loop) \\\hline
    32×32 RGB Images & 32×32 RGB Images & 32×32 RGB Images\\\hline
    3$\times$3 Conv2d, ReLU & 3$\times$3 Conv2d, ReLU& 5$\times$5 Conv2d, ReLU\\
    3$\times$3 Conv2d, ReLU & 3$\times$3 Conv2d, ReLU & 2$\times$2 Max-pool \\
    2$\times$2 Max-pool & 2$\times$2 Max-pool & \\\hline
    3$\times$3 Conv2d, ReLU & 3$\times$3 Conv2d, ReLU & 3$\times$3 Conv2d, ReLU\\
    2$\times$2 Max-pool & 3$\times$3 Conv2d, ReLU & 2$\times$2 Max-pool\\
    & 2$\times$2 Max-pool & \\\hline
    Dense 8192$\rightarrow$1024, ReLU& 3$\times$3 Conv2d, ReLU & 3$\times$3 Conv2d, ReLU\\
    Dense 1024$\rightarrow$256, ReLU & 3$\times$3 Conv2d, ReLU & 2$\times$2 Max-pool\\
    & 2$\times$2 Max-pool & \\\hline
    \multirow{3}{*}{Dense 256$\rightarrow$10}&Dense 2048$\rightarrow$1024, ReLU& Dense 512$\rightarrow$64\\
    & Dense 1024$\rightarrow$512, ReLU & Dense 64$\rightarrow$10\\
    & Dense 512$\rightarrow$10\\\hline
    \end{tabular}
    \vspace{3pt}
    \caption{The network structures of the models used in our experiments.}
    \label{tab:network_structures}
\end{table}

\section{Supplementary Experimental Results}\label{supp:results}

\begin{table}[!t]
    \centering
    \begin{tabular}{l|cccc}
    \toprule
        Imperfect supervision & $k=1000$ & $k=2000$ & $k=3000$ & $k=4000$\\\midrule
        With 30\% corrupted labels & 951.2$\pm$4.9  & 1866.1$\pm$8.3 & 2713.7$\pm$10.8 & 3675.6$\pm$17.0\\\midrule
        With 50\% corrupted labels & 934.5$\pm$5.6 & 1856.5$\pm$9.1 & 2708.8$\pm$11.2 & 3668.4$\pm$14.6 \\\midrule
        With class-imbalanced data & 988.4$\pm$6.7 & 1893.8$\pm$10.0 & 2762.7$\pm$14.2 & 3757.4$\pm$17.8\\
       \bottomrule
    \end{tabular}
    \vspace{3pt}
    \caption{Mean and standard deviation of optimized coreset sizes by our method under imperfect supervision.}
    \label{tab:imperfect}
\end{table}

\subsection{The Average Accuracy Brought by Per Data Point}\label{supp:mean_per_data_point}
\begin{figure}[!t]
    \centering
    \includegraphics[width=0.55\textwidth]{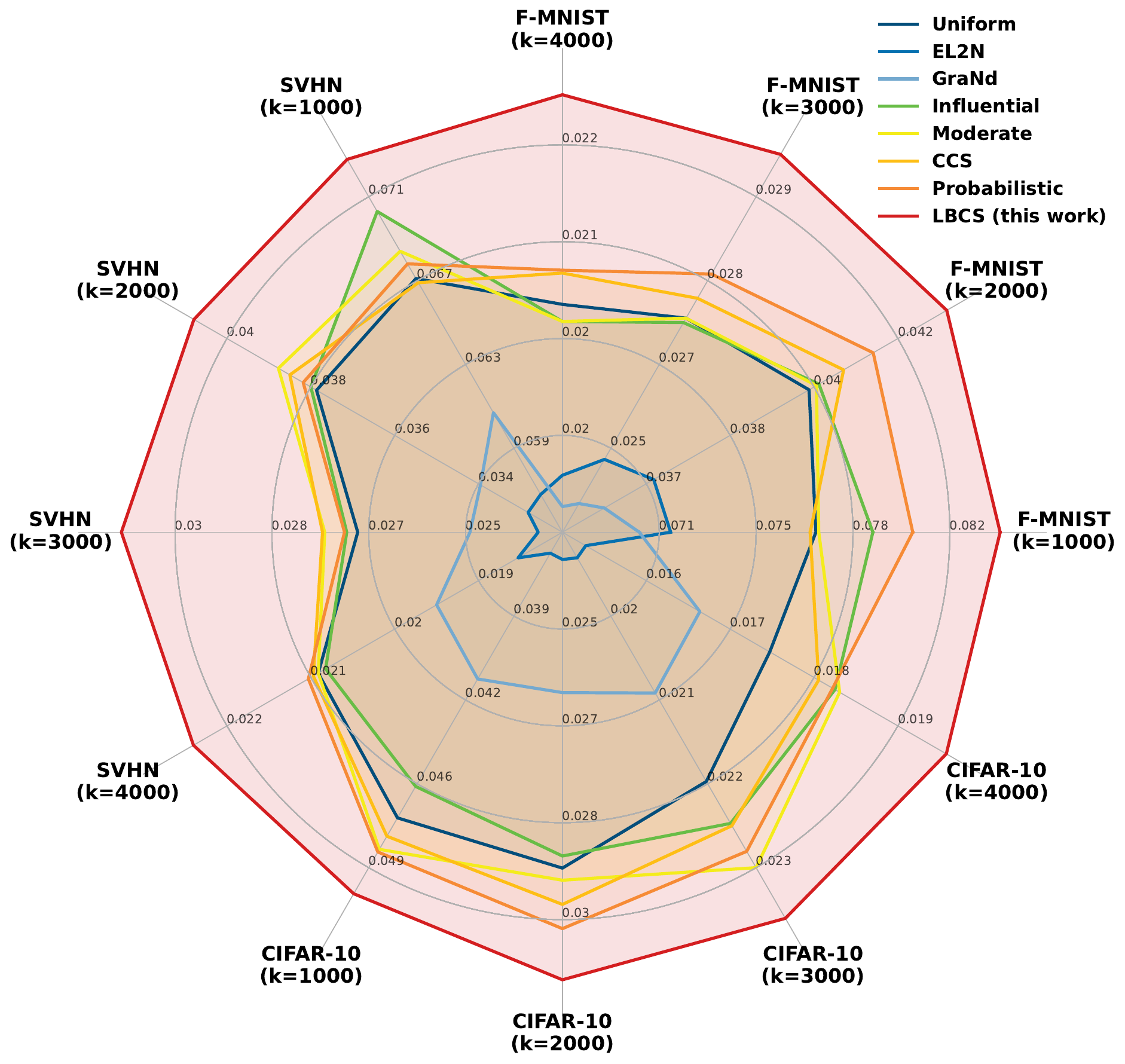}
    \caption{The illustration of the average accuracy~(\%) brought by \textit{per data point} within the selected coreset.}
    \label{fig:acc_per_data}
\end{figure}

In Figure~\ref{fig:acc_per_data}, we report the average accuracy brought by per data point within the selected coreset. As can be seen, the proposed LBCS
always enjoys higher average accuracy.

\begin{figure}[!t]
    \centering
    \includegraphics[width=0.55\textwidth]{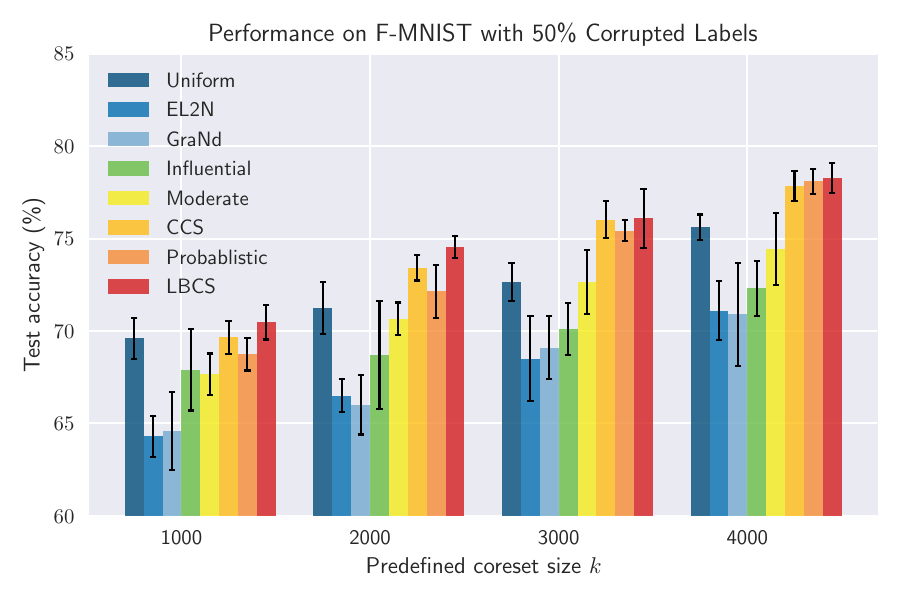} 
 \vspace{-7pt}
    \caption{Illustrations of coreset selection with with 50\% corrupted labels. The optimized coreset size by LBCS is provided in Appendix~\ref{supp:im_coreset_size}.} 
    \label{fig:im_b}
\end{figure}

\subsection{Results with 50\% Label Noise}\label{supp:50_noise}
In the main paper, we evaluate the effectiveness of the proposed method when the noise rate is 30\%. Here, we consider a more challenging case, where the noise rate is increased to 50\%. Experimental results are provided in Figure~\ref{fig:im_b}. As can be seen, even though the noise level is high, the proposed method still achieves the best performance.

\subsection{Optimized Coreset Sizes with Imperfect Supervision}\label{supp:im_coreset_size}
In the main paper~(\S\ref{sec:imperfect}), we have shown the strength of the proposed method in coreset selection under imperfect supervision. Here we supplement the optimized coreset sizes by our method in these cases, which are provided in Table~\ref{tab:imperfect}.

\subsection{Ablation on Search Times}\label{supp:search_time}
We provide the ablation study on the search time $T$ in Table~\ref{tab:acc_size}. 
Initially, as the search times increase, there is a noticeable upward trend in test accuracy accompanied by a decrease in coreset size. As the search progresses, the test accuracy gradually stabilizes, and the coreset size maintains a consistently smaller value. Subsequently, with a large number of searches, the search results exhibit limited changes, indicating empirical convergence in the search process. In practical terms, selecting an appropriate value for $T$ can be tailored to specific requirements for coresets and the allocated budget for coreset selection.

\begin{table}[!t]
    \centering
    \begin{tabular}{c|c|c||c|c}
    \toprule 
    \multirow{2}{*}{$T$}    & \multicolumn{2}{c||}{$k=1000$} & \multicolumn{2}{c}{$k=2000$} \\\cmidrule{2-5}
    & Test acc. & Coreset size (ours) & Test acc. & Coreset size (ours) \\\midrule
    100 &  77.0$\pm$1.8 &  998.0$\pm$1.9 & 80.2$\pm$1.9 & 1995.6$\pm$2.5\\
    200 & 77.7$\pm$1.5 & 990.3$\pm$2.3 & 80.9$\pm$1.0 & 1976.3$\pm$4.7\\
    300 & 78.5$\pm$1.2 & 975.6$\pm$2.7 & 81.7$\pm$0.7 & 1945.5$\pm$3.9\\
    500 & 79.7$\pm$0.7 & 956.7$\pm$3.5 & 82.8$\pm$0.6 & 1915.3$\pm$6.6\\
    800 & 79.2$\pm$0.8 & 940.7$\pm$4.7 & 82.5$\pm$0.5 & 1905.7$\pm$5.4\\
    1500 & 79.5$\pm$0.5 & 935.4$\pm$4.9 & 82.7$\pm$0.6 & 1894.1$\pm$4.1\\
    2000 & 79.8$\pm$0.6 & 935.8$\pm$3.8 & 82.8$\pm$0.8 & 1893.9$\pm$4.3\\ \bottomrule
    \end{tabular}
    \caption{Ablation study of the number of search times.}
    \label{tab:acc_size}
\end{table}

\subsection{Evaluations with Different Network Architectures}\label{supp:network_results}

Here we demonstrate that the proposed method is not limited to specific network architectures. We employ SVHN and use ViT-small~\citep{dosovitskiy2021image} and WideResNet~(abbreviated as W-NET)~\citep{zagoruyko2016wide} for training on the constructed coreset. The other experimental settings are not changed. Results are provided in Table~\ref{tab:exact_results_network}. As can be seen, with ViT, our method is still superior to the competitors with respect to test accuracy and coreset sizes~(the exact coreset sizes of our method can be checked in Table~\ref{tab:exact_results}). With W-NET, our LBCS gets the best test accuracy when $k=1000$, $k=3000$, and $k=4000$ with smaller coreset sizes. In other cases, \ie, $k=2000$, LBCS can achieve competitive test accuracy compared with baselines but with a smaller coreset size.

\begin{table}[!t]
    \centering
    \setlength{\tabcolsep}{1.3mm}{
    \begin{tabular}{l|c|ccccccc|>{\columncolor{Gray}}c}
    \toprule
     & $k$ &Uniform & EL2N & GraNd &Influential & Moderate &  CCS & Probablistic & LBCS~(ours)\\\midrule
    \parbox[t]{2mm}{\multirow{4}{*}{\rotatebox{90}{\tiny ViT}}}& 1000 & 28.5$\pm$3.1 & 22.7$\pm$3.5 & 24.0$\pm$2.2 & 31.5$\pm$1.8 & 32.8$\pm$1.5 & 31.7$\pm$1.6 & 29.6$\pm$0.3 & \textbf{33.9$\pm$0.8}\\
    &2000 & 46.6$\pm$2.7 & 40.9$\pm$2.6 & 38.8$\pm$0.6 & 42.2$\pm$1.7 & 45.5$\pm$2.3 & 46.1$\pm$1.8 & 46.6$\pm$2.0 & \textbf{47.5$\pm$2.2}\\
    &3000 & 50.0$\pm$2.2 & 46.7$\pm$3.0 & 47.9$\pm$2.4&50.8$\pm$0.7 & 51.0$\pm$2.9 & 50.4$\pm$1.6 & 50.5$\pm$1.9 & \textbf{51.3$\pm$0.6}\\
    &4000 & 54.0$\pm$3.3 & 49.9$\pm$2.8 & 50.8$\pm$0.9 & 53.3$\pm$0.9 & 54.9$\pm$1.9 & 56.2$\pm$2.1 & 55.3$\pm$1.5 & \textbf{57.7$\pm$0.4}\\\midrule
    \parbox[t]{2mm}{\multirow{4}{*}{\rotatebox{90}{\tiny W-NET}}}& 1000 & 78.8$\pm$1.5 & 67.9$\pm$2.7 & 70.5$\pm$3.0 & 79.3$\pm$2.8 & 80.0$\pm$0.4 & 79.8$\pm$0.9 & 80.1$\pm$1.3 & \textbf{80.3$\pm$1.2}\\
    &2000 & 87.2$\pm$1.2 & 69.5$\pm$3.3 & 73.4$\pm$2.6 & 87.1$\pm$0.8 & 88.0$\pm$0.3 & \textbf{88.7$\pm$0.6} & 87.0$\pm$1.0 & 87.8$\pm$1.1\\
    &3000 & 89.1$\pm$0.9 & 76.6$\pm$1.2 & 78.8$\pm$3.2 & 90.3$\pm$0.7 & 90.3$\pm$0.4 &90.2$\pm$0.4 & 89.3$\pm$0.6 & \textbf{90.7$\pm$0.5}\\
    &4000 & 90.2$\pm$1.9 & 80.3$\pm$1.9 & 83.4$\pm$1.7 & 90.9$\pm$1.1 & 90.8$\pm$0.5 & 91.1$\pm$1.0 & 90.6$\pm$0.5 & \textbf{91.4$\pm$0.9}\\
    \bottomrule
    \end{tabular}}
    \caption{Mean and standard deviation (std.) of test accuracy (\%) on SVHN with various predefined coreset sizes and networks. The best mean test accuracy in each case is in \textbf{bold}. }
    \label{tab:exact_results_network}
\end{table}

\subsection{Setups and Results of Continual Learning with Constructed Coresets}\label{supp:cl_results}
Continual learning targets non-stationary or changing environments, where a set of tasks needs to be completed sequentially~\citep{wang2023comprehensive,kim2022a,peng2022continual}. The constructed coresets can be used to keep a subset of data related to previous tasks, which alleviates the catastrophic forgetting of early knowledge. 

For experimental setups, as did in~\citep{borsos2020coresets}, we exploit PermMNIST~\citep{goodfellow2013empirical}, which consists of 10 tasks, where in each task the pixels of all images undergo the same fixed random permutation. The memory size is set to 100. As previous work~\citep{zhou2022probabilistic} did not provide the code about this part, we employ the implementation of~\citep{borsos2020coresets} for continual learning with coresets\footnote{\href{https://github.com/zalanborsos/bilevel_coresets}{https://github.com/zalanborsos/bilevel\_coresets}}. The weight for previous memory is set to 0.01. In addition, we inject 10\% symmetric label noise into training data to evaluate the robustness of our method in this case. We provide the experimental results in Table~\ref{tab:cl}. As can be seen, our LBCS consistently works better than all baselines.


\begin{table}[!t]
    \centering
    \begin{tabular}{l|ccccccc|>{\columncolor{Gray}}c}
    \toprule
    Method  &  Uniform & EL2N & GraNd & Influential & Moderate & CCS & Probabilistic & LBCS\\\midrule
    PermMNIST  & 78.1 & 75.9 & 77.3 & 78.8 & 78.4 & 79.4 & 79.3 & \textbf{79.9}\\\midrule
    Noisy PermMNIST & 65.0 & 52.8 & 61.8 & 64.0 & 63.9 & 64.6 & 65.5 & \textbf{65.9}\\
    \bottomrule
    \end{tabular}
    \vspace{3pt}
    \caption{Experimental results of \textit{continual learning with constructed coresets}. ``Noisy PermMNIST'' indicates the label-noise version of PermMNIST. The best result in each case is in \textbf{bold}.}
    \label{tab:cl}
\end{table}

\begin{table}[!t]
    \centering
    \begin{tabular}{l|ccccccc|>{\columncolor{Gray}}c}
    \toprule
    Method  &  Uniform & EL2N & GraNd & Influential & Moderate & CCS & Probabilistic & LBCS\\\midrule
    PermMNIST  & 69.0 & 70.2 & 71.7 & 70.9 & 68.0 & 71.7 & 72.1 & \textbf{73.2}\\\midrule
    Noisy PermMNIST & 59.4 & 58.4 & 63.3 & 62.4 & 60.0 & 65.3 & 64.6 &\textbf{66.1}\\
    \bottomrule
    \end{tabular}
    \vspace{3pt}
    \caption{Experimental results of \textit{streaming with constructed coresets}. ``Noisy PermMNIST'' indicates the label-noise version of PermMNIST. The best result in each case is in \textbf{bold}.}
    \label{tab:s}
\end{table}

\subsection{Setups and Results of Streaming with Constructed Coresets}\label{supp:streaming_results}
Streaming is similar to continual learning but is more challenging~\citep{aljundi2019gradient,hayes2019memory,chrysakis2020online}. In streaming, there is no explicit concept of tasks. Data is sequentially provided to the model. In these circumstances, coresets can be employed to build the replay memory, where selected data points represent each task. 

For experimental setups about streaming with constructed coresets, we follow the implementation in~\citep{borsos2020coresets}. For this experiment, we modify PermMNIST by first concatenating all tasks for the dataset and then streaming them in batches of size 125. The replay memory size and the number of slots are set to 100 and 0.0005 respectively. Networks are trained for 40 gradient descent steps using Adam with step size 0.0005 after each batch. We provide the experimental results in Table~\ref{tab:s}, which demonstrates the effectiveness of our method in streaming with built coresets. Note that the results are somewhat different from the report in~\citep{zhou2022probabilistic}. It is because \citep{zhou2022probabilistic} did not provide the code of streaming with constructed coresets in the GitHub repository. We hence use the implementation of \citep{borsos2020coresets}.


\section{More Related Work}\label{supp:related_work}
\subsection{Data Distillation}
Data distillation~\citep{wang2018datasetdistillation,lee2022dataset,wang2022cafe,deng2022remember,loo2022efficient,zhang2023accelerating} is an alternative approach for dataset compression, which is inspired by knowledge distillation. Different from coreset selection, this series of works focuses on \textit{synthesizing} a small but informative dataset as an alternative of the original dataset. However, data distillation is criticized for only synthesizing a small number of data points~(\eg, 1/10 images per class) due to computational source limitations~\citep{yang2023dataset}. Its performance is far from satisfactory. In addition, from the perspective of human perception, the distillation often destroys the semantic information of original data. Therefore, this paper is consistent with previous works~\citep{zhou2022probabilistic,sorscher2022beyond,yang2023dataset,xia2023moderate}. The performances of data distillation and coreset selection are not compared directly.

\section{Reproducibility}\label{supp:repro}
In the reviewing process, we anonymously provide the source code in the supplementary material. Also, the code will be made public after paper acceptance.

\section{Limitations}\label{supp:impacts_limitations}
The proposed method is based on bilevel optimization coreset selection. At present, there are also some advanced methods that do not need bilevel optimization. This work does not discuss an effective way to involve the minimization of the coreset size in those methods. Also, although theoretical analysis provides convergence guarantees, the optimal convergence rate remains mysterious. We regard addressing the limitations as future research directions. 
\end{document}